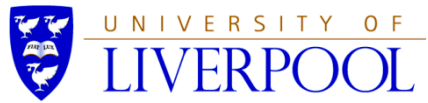

# Developing a supervised training algorithm for limited precision feed-forward spiking neural networks

By

# Evangelos Stromatias

Dissertation submitted to the University of Liverpool in partial fulfillment of the requirements for the degree of Master of Science in Engineering in Microelectronic Systems.

September, 2011

# Acknowledgements

First of all I would like to thank my parents since without their support I would not be able to do this Master program. I would like to express my gratitude to my supervisor Dr. John Marsland for proposing and for giving me the opportunity to do a project on spiking neural networks and also for his guidance and support throughout my project. Furthermore, I would like to thank my assessor, Dr. Siavash Amin-Nejad for his help during the writing of this thesis. In addition, I would like to thank Gordon and Jeff for their help through the year and all of my friends for being there for me. Nonetheless, I would mostly like to thank the following people for their help and inspiration: Dr. Grigoris Nikolaou, Marios Daoutis and Athanasios Charalampopoulos. Finally, I would like to thank Thaleia Georgiou for her love and encouragement.

**Nothing is particularly hard if you divide it into small jobs.**

Henry Ford

*U.S. automobile industrialist (1863 - 1947)*

# Abstract

**Developing a supervised training algorithm for limited precision feed-forward spiking neural networks**

**Proposed by: Dr. John Marsland**

Spiking neural networks have been referred to as the third generation of artificial neural networks where the information is coded as time of the spikes. There are a number of different spiking neuron models available and they are categorized based on their level of abstraction. In addition, there are two known learning methods, unsupervised and supervised learning. This thesis focuses on supervised learning where a new algorithm is proposed, based on genetic algorithms. The proposed algorithm is able to train both synaptic weights and delays and also allow each neuron to emit multiple spikes thus taking full advantage of the spatial-temporal coding power of the spiking neurons. In addition, limited synaptic precision is applied; only six bits are used to describe and train a synapse, three bits for the weights and three bits for the delays. Two limited precision schemes are investigated. The proposed algorithm is tested on the XOR classification problem where it produces better results for even smaller network architectures than the proposed ones. Furthermore, the algorithm is benchmarked on the Fisher iris classification problem where it produces higher classification accuracies compared to SpikeProp, QuickProp and Rprop. Finally, a hardware implementation on a microcontroller is done for the XOR problem as a proof of concept.

**Table of Contents**

# List of Figures

# List of Tables

# 1. Introduction

**1.1. The three generations of artificial neural networks**

Artificial neural networks are an important part of artificial intelligence and have been extensively used in pattern recognition, medical diagnosis, image analysis, finance, weather prediction and many more computer science and engineering tasks. They are massively parallel-distributed processors in contrast to the conventional computers and are able to learn patterns through synaptic plasticity. Wolfgang in one of his published works [1] divided the artificial neural networks into three generations based on how biologically close to the real neurons are. The following section summarizes these three generations.

The first generation of the artificial neural networks used the McCulloch-Pitts (1943) threshold neuron model. This neuron model, also known as perceptron, has two states: 'High' or 'Low' based on the sum of the input signals multiplied by its weights. This type of neural network is able to compute all Boolean functions.

The 2$^{nd}$ generation of artificial neural networks used a sigmoid activation function instead of a threshold. By using a continuous activation function these neurons could be used for analog computations. Furthermore, they are more powerful than the neurons of the 1$^{st}$ generation, since for digital computations they require fewer neurons [2]. In addition, a new supervised learning algorithm could be used, the error backward propagation based on the gradient descent, which changes the values of the weights in order to minimize the output error. From a biological point of view the 2$^{nd}$ generation artificial neurons are more realistic than the 1$^{st}$ generation ones because the output of a sigmoidal unit can be looked as the firing rate of a biological neuron [1].

Yet real neurons in the cortex can perform fast analog computations, like facial recognition that takes 100ms, which means that the processing time per neuron could not be more than 10ms [3]. These results show that the time window is too small for rate coding. However, that does not mean that rate coding is not biologically meaningful; it has been proven experimentally, in the case of touch receptor in the leech [4], where more spikes produced stronger touch, during a stimulating period.

Researches have shown that neurons communicate with spikes, also known as action potentials. Since all spikes are identical: 1-2ms of duration and 100mV of amplitude [5], the information is encoded by the timing of the spikes and not the spikes themselves. These experimental results gave birth to the 3rd generation of artificial neural networks, the spiking neural networks that are more biologically close to the real neurons. They can encode temporal information in their signals, which means that they can incorporate spatial-temporal information. Wolfgang [6], in one of his works, showed that it is possible to simulate sigmoidal gates with the temporal information of the spikes.

**1.2. The biological neuron**

A typical neuron is divided into three parts: the dendrites, the soma and the axon. Generally speaking, the dendrites are receiving the input signals from the previous neurons. The soma is where the received input signals are being processed and the axon is where the output signals are transmitted. The synapse is between every two neurons; if a neuron j sends a signal across the synapse to neuron i, the neuron that sends the signal is called presynaptic and the neuron that receives the signal is called postsynaptic neuron.

Hodgkin and Huxley [7] found out, by experimenting on the squid giant axon, that it is the time of the spikes that encodes information [8], Figure 1.1.

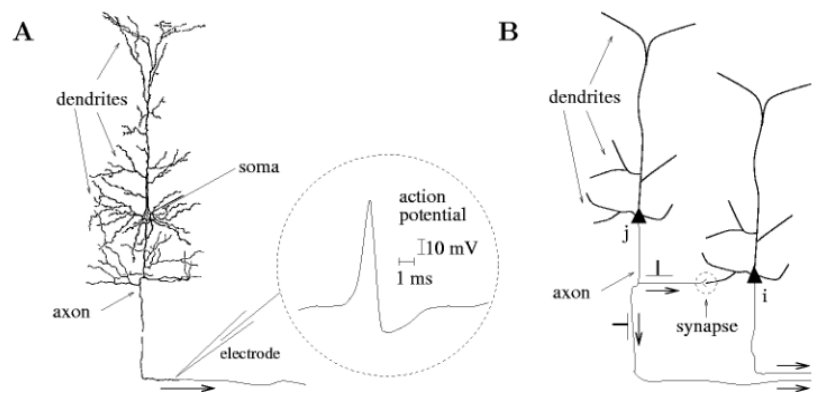

Figure 1.1: A. The inset shows an example of a neuronal action potential. The action potential is a short voltage pulse of 1-2ms duration and 100mV of amplituted. B. Signal transmition from a presynaptic neuron j to a post synaptic neuron i. The synapse is marked by a dashed circle [5].

### 1.2.1 The membrane potential

Every neuron is surrounded by positive and negative ions. In the inner surface of the membrane there is an excess of negative charges and on the outer surface there is an excess of positive charges. Those charges create the membrane potential.

The membrane potential can be calculated from the following equation: Vm=Vin-Vout, where Vin is the negative charges on the inside of the cell and Vout are the positive charges outside of the cell. When the membrane potential is at the resting state, that is when it is not receiving any input signals, the resting potential Vrest is set to Vin, which is around -60mV to -70mV.

When the neuron receives an input, some of the ion channels of the cell open and others close, resulting in an electrical current flow into the cell, which results in a change of the resting potential Vrest [4].

The phenomenon during which the membrane's potential changes exceed the resting potential is called depolarization. The opposite phenomenon is called hyperpolarization. When the depolarization reaches a critical value, also known as threshold, the cell produces an action potential (a spike) [4], figure 1.2. If the membrane potential receives an input that causes depolarization or hyperpolarization and after that does not receive any other input, the membrane potential returns slowly to its resting potential.

In the case of the Glial cell the potassium K+ are flowing from the inside of the cell to the outside causing a potential difference called equilibrium potential $E_k$ [4] . This $E_k$ determines the resting membrane potential and can be calculated from the Nerst Equation:

$$E_k = \frac{RT}{zF} \ln \frac{[X]_o}{[X]_i} \quad (1.1)$$

Where R is the gas constant, T is the temperature in Kelvin, z is the valence of the ion, F the Faraday constant, [X]o and [X]i are the concentrations of the ion outside and inside of the cell [4]. Thus the Vrest for the Glial cell is Vrest= -75mV. The membrane potential will be discussed in the next section when the Hodgkin-Huxley neuron model will be described based on the experiments on the squid giant axon.

**1.2.2 The action potential**

As stated before, when the membrane potential reaches a critical value called threshold it emits an action potential, also known as a spike. This is caused by the movement of ions across the membrane through voltage-gated channels [4]. The spikes are identical to each other and their form does not change as the signal moves from a presynaptic to a postsynaptic neuron [5]. The firing times of a neuron are called spike train and it is represented with the following equation:

$$Fi = \{t_i^1, t_i^2, ..., t_i^{(n)}\} \quad (1.2)$$

The subscript i defines the neuron and the superscript defines the number of the emitted spikes, where (n) is the most recent emitted spike.

Directly after the transmission of a spike, the membrane potential goes through a phase of high hyperpolarization under the resting potential and then slowly returns back to the resting potential. During that time, it is not possible to emit a second spike even for strong input signals, that is because the ion channels are open instantly after a spike has been generated [5]. The minimum time between two generated spikes is called absolute refractory period and the phenomenon where the membrane potential undershoots below the resting potential is known as the spike after potential (SAP), Figure 1.2.

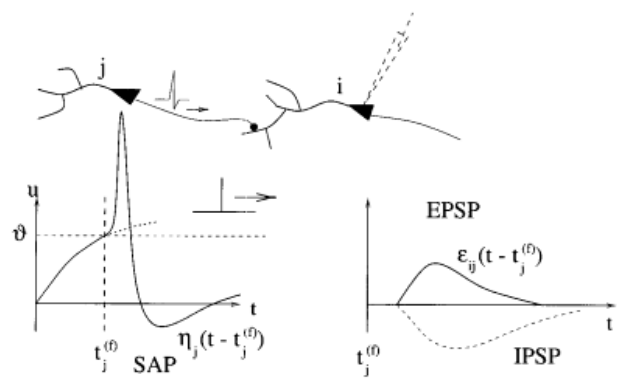

**Figure 1.2:** The membrane potential is increased and at time tj(f) the membrane potential reaches the threshold so a spike is emmited [8].

**1.2.3 The synapse**

Between the axon of the presynaptic neuron and the dendrite of the postsynaptic neuron there is a small gap, also known as synaptic gap. The operation of the synapse is very complicated and a detailed description is beyond the scope of this review.

The spike of the presynaptic neuron cannot cross this gap, however, when a spike arrives from the presynaptic neuron to the synapse the gap is filled a fluid that generates a postsynaptic potential (PSP) to the dendrite of the postsynaptic neuron [4]. This process does not happen instantaneous; there is a small delay generated in that particular synapse.

There are two types of postsynaptic potentials. If the generated postsynaptic potential is positive it is called excitatory postsynaptic potential (EPSP) or if the generated postsynaptic potential is negative it is call inhibitory postsynaptic potential (IPSP), Figure 1.3. An IPSP lowers the membrane potential of the postsynaptic neuron while an EPSP increases it and may cause it to fire a spike.

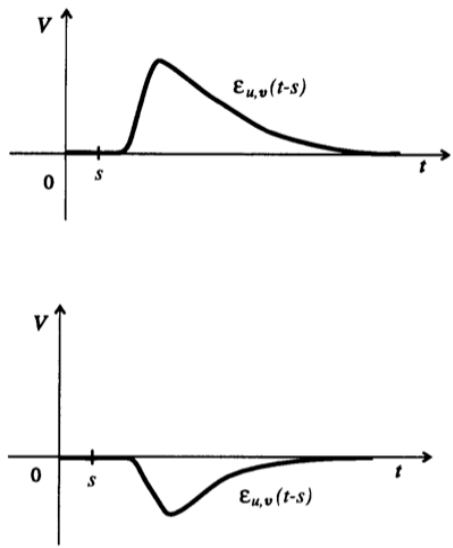

**Figure 1.3: Excitatory postsynaptic potential (EPSP) and Inhibitory postsynaptic potential (IPSP) of a biological neuron [1].**

The weight of a synapse $w^k_{ji}$ and the axonal delay $d^k$ are used as parameters for training a neural network, where k is the synapse between a presynaptic neuron i to a postsynaptic neuron j. Recent experimental results have shown that between every two interconnected neurons can be multiple synapses with multiple delay times [9,10], Figure 1.4. Those delays have been used in a numerous works as training parameters [11-14].

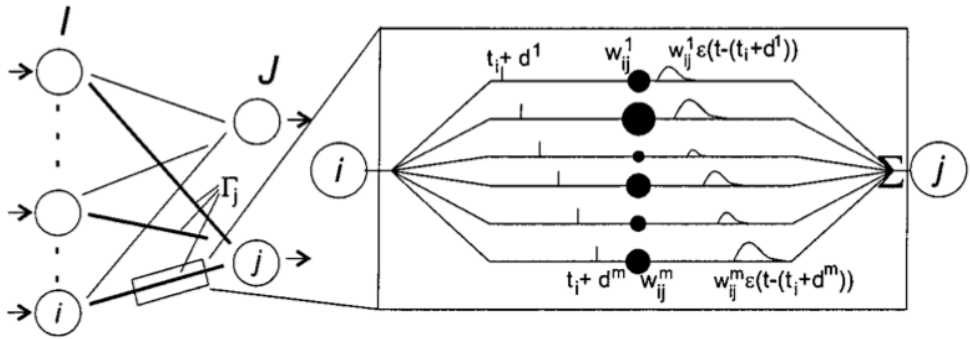

Figure 1.4: Neurons in layer J (postsynaptic) receive connections from neurons Γj (presynaptic) in layer I. Inset: a single connection between 2 neurons consists of m delayed synaptic terminals. A synaptic terminal k is associated with a weight $w^k_{ji}$ and delay $d^k$ [11]. The black circles represent the magnitude of the synaptic weight.

## 1.3. Spiking Neuron Models

Spiking neuron models can be divided into two major categories [5] based on their level of abstraction: The conductance models and the threshold models. The conductance models simulate the ion channels of the cell, while the threshold models represent a higher level of abstraction where the threshold voltage has a fixed value and the neuron fires every time the membrane potential reaches it.

There are two additional models that will not be described in this thesis: the compartmental and rate models. The compartmental models will not be discussed due to their complexity and the rate models are actually the sigmoidal neurons that are used in the traditional artificial neural networks of the 2$^{nd}$ generation. Due to their nature, they neglect all the temporal information of the spikes and only describe their activity as spike rates.

### 1.3.1 Conductance-Based Models

In general, Conductance-Based models have been derived from the Nobel prize winners (1963) Hodgkin and Huxley [8], based on the experiments that they performed on the giant axon squid [7]. Basically, they describe what happens to the ion channels of the neuron cell.

### *1.3.1.1. Hodgkin-Huxley Model*

The schematic diagram of the Hodgkin and Huxley model can be seen in Figure 1.5. The semipermeable cell membrane that separates the interior of the cell from the exteracellular liquid acts as a capacitor [5]. The input signal is the current I(t) and the batteries represent the Nerst potential generated by the difference in ion concentration. This model has three types of ionic current: sodium (Na), potassium (K) and a leak current.

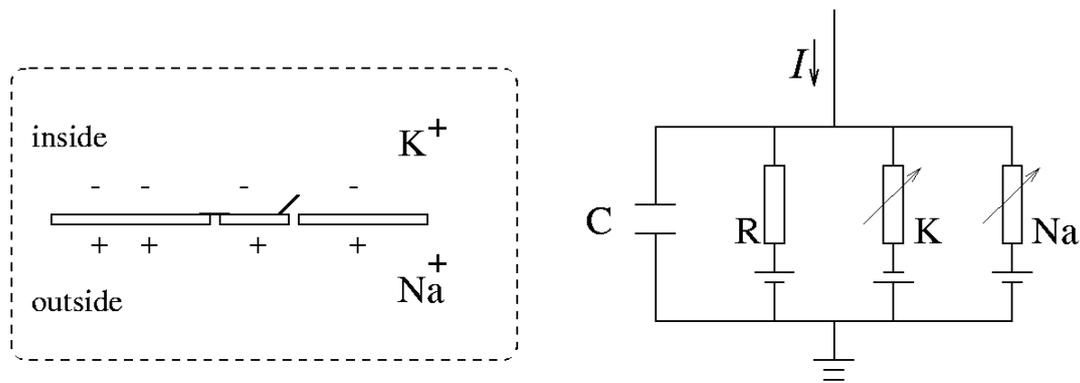

**Figure 1.5: Schematic diagram of the Hodgkin-Huxley model [5].**

The model can be described as an RC circuit; equation 1.3 is the mathematical expression of Figure 1.5, where $I_k$ are the sums of all ion channels: a sodium channel Na, a potassium channel K and an unspecific leakage channel with R resistance. The membrane capacity is C= 1µF/cm².

$$I(t) = I_c(t) + \sum_k I_k(t) \quad (1.3)$$

In order to be able to compute the membrane potential at any time t the equation 1.3 becomes:

$$C\frac{du}{dt} = -\sum_k I_k(t) + I(t) \quad (1.4)$$

In equation 1.5 the sum of the I$_k$ can be seen. It is worth mentioning that if all ion channels are open, they transmit currents with a maximum conductance g$_{NA}$ or g$_K$. However, this cannot happen since several of the channels are closed. The additional variables *m*, *n* and *h*, also known as gating variables, describe the probability of a channel being open. The *m* and *h* control the Na⁺ channels and the *n* control the K⁺ channels.

$$\sum_k I_k = g_{Na} m^3 h (u - E_{Na}) + g_k n^4 (u - E_K) + g_L (u - E_L) \quad (1.5)$$

Where E$_{NA}$, E$_K$ and E$_L$ are parameters and represent the reversal potentials. The variables *m*, *n* and *h* are described by the following differential equations:

$$m' = a_m(u)(1-m) - b_m(u)m \quad (1.6)$$

$$n' = a_n(u)(1-n) - b_n(u)n \quad (1.7)$$

$$h' = a_h(u)(1-h) - b_h(u)h \quad (1.8)$$

The functions α and β are given in the Table 1.1; they are empirical functions of u that have been created by Hodgin and Huxley to fit the experimental data of the giant axon of the squid.

Table 1.1: Parameters of the Hodgkin-Huxley equations [5].

| $x$ | $E_x$ | $g_x$ |
|---|---|---|
| Na | 115 mV | 120 mS/cm$^2$ |
| K | -12 mV | 36 mS/cm$^2$ |
| L | 10.6 mV | 0.3 mS/cm$^2$ |

| $x$ | $\alpha_x(u\,/\,\mathrm{mV})$ | $\beta_x(u\,/\,\mathrm{mV})$ |
|---|---|---|
| $n$ | $(0.1 - 0.01\,u) / [\exp(1 - 0.1\,u) - 1]$ | $0.125 \exp(-u\,/\,80)$ |
| $m$ | $(2.5 - 0.1\,u) / [\exp(2.5 - 0.1\,u) - 1]$ | $4 \exp(-u\,/\,18)$ |
| $h$ | $0.07 \exp(-u\,/\,20)$ | $1 / [\exp(3 - 0.1\,u) + 1]$ |

If some external input causes the membrane voltage to rise, the conductance of the sodium (Na) channels increases due to increasing *m*. If this positive feedback is large enough, an action potential (spike) is initiated.

At high values of u the sodium (Na) conductance is shut off due to factor *h*. Thus the variable *h*, which closes the channels, reacts more slowly to the voltage increase than the variable *m*, which opens the channel.

The potassium (K) current lowers the potential because it has an outward direction; potassium current is controlled by *n*. The overall effect of the sodium and potassium currents is a short action potential (spike) followed by a negative overshoot.

If we apply the same input current $I_{(t)}$ shortly after an action potential, it is almost impossible to achieve a second action potential due to the absolute *refractoriness*, Figure 1.6. Finally, if the value of $I_{(t)}$ is less than a critical value $I_\theta=6\mu A/cm^2$, the membrane potential returns to the rest value without causing an action potential.

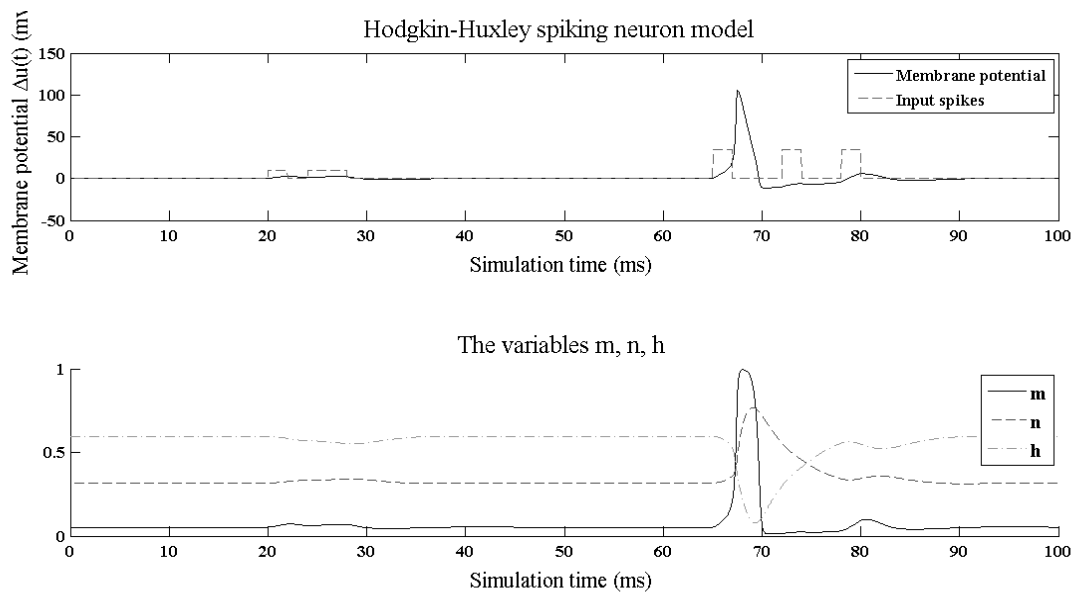

**Figure 1.6: The neuron at first it receives 2 input pulses of $2\mu A/cm^2$ amplitude but they are not strong enough to emit a spike. At t=65 ms the neuron receives an input spike of $7\mu A/cm^2$ amplitude that causes an action potential (spike), at time t=72ms and t=78 it receives the same input pulses as the one that caused the spike but it does not fire due to refractoriness (notice the gating variables m,n,h). Appendix A.1.**

***1.3.1.2. The Izhikevich spiking neuron model***

Izhikevich [16] created a model, which combines the dynamics of the Hodgkin-Huxley model and the computational efficiency of the threshold-fire models. This was done by reducing the 4 dimensional model of the Hodgkin-Huxley model into two first order differential equations, equations (1.9) and (1.10).

$$v' = 0.04v^2 + 5v + 140 - u + I \quad (1.9)$$

$$u' = a(bv - u) \quad (1.10)$$

The variable **v** represents the membrane potential of the neuron and **u** represents a membrane recovery variable, which is the activation of potassium K ionic currents and inactivation of sodium Na ionic currents. This model can exhibit all known neuronal firing patterns with the appropriate values for the a, b, c and d variables. Furthermore, this model has a dynamic threshold that depends on the previous state of the membrane potential before the spike. In the next section the parameters of equation (1.10) are explained [16].

- The parameter **a** describes the time scale of the recovery variable u. Smaller values result in slower recovery. A typical value is a = 0.02.
- The parameter **b** describes the sensitivity of the recovery variable u to the sub-threshold fluctuations of the membrane potential v. A typical value is b=0.2.
- The parameter **c** describes the after-spike reset value of the membrane potential v caused by the fast high-threshold K (potassium) conductance. A typical value for real neurons is c=-65mV.
- The parameter **d** describes the after-spike reset of the recovery variable u caused by slow high threshold Na (sodium) and K (potassium) conductance. A typical value is d=2.

In Figure 1.7 the parameters a, b, c and d of the neuron model can be observed and in Figure 1.8 the membrane potential of the Izhikevich neuron model can be seen.

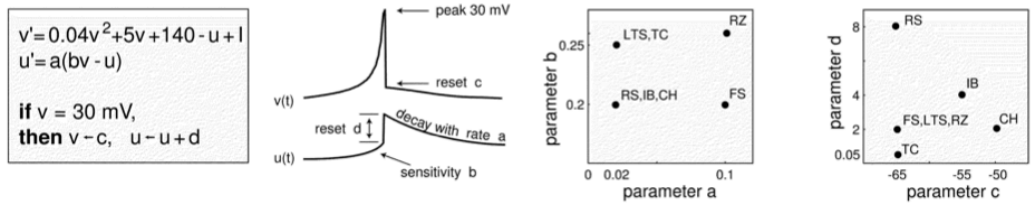

Figure 1.7: Parameters of the spiking neuron model [16].

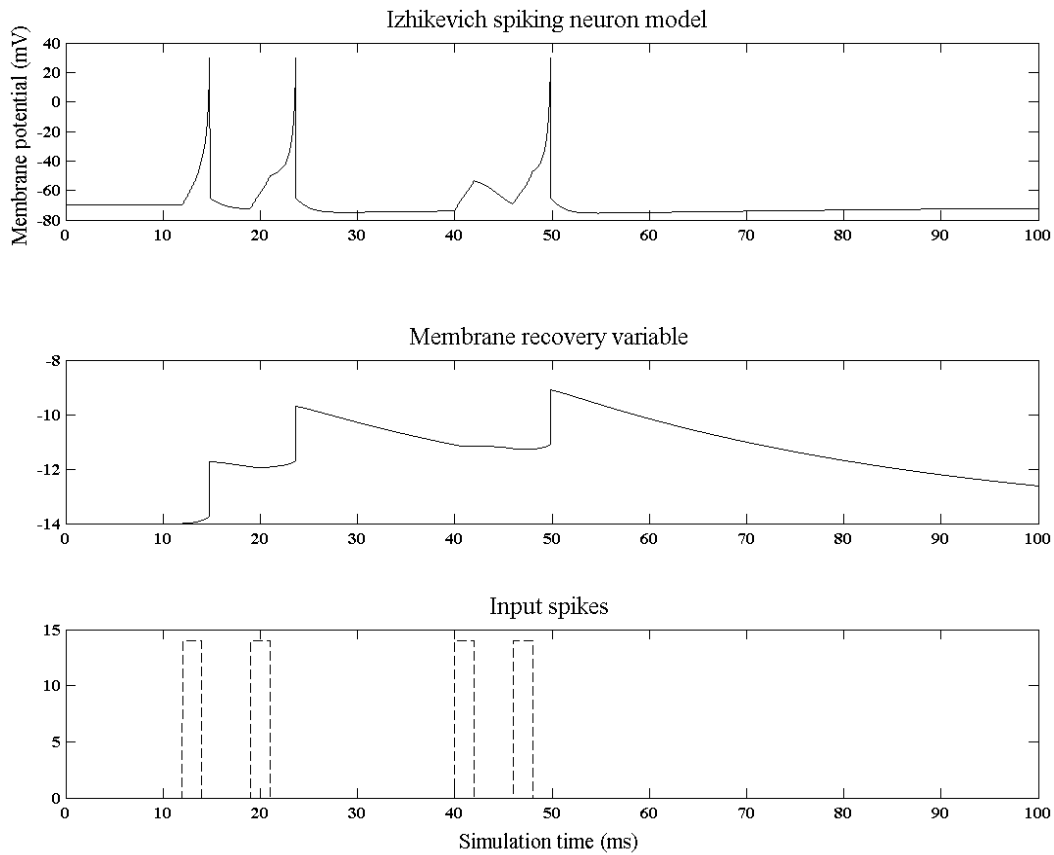

Figure 1.8: The Izhikevich Spiking Neuron Model. In the top graph is the membrane potential of the neuron. In the middle graph is the membrane recovery variable. Finally the bottom plot represents the action presynaptic spikes. Appendix A.2.

### 1.3.2. Threshold-Fire Models

The threshold-fire models represent a higher level of abstraction compared to the conductance-based models. These models are based on the summation of all contributions of the presynaptic neurons to the membrane potential. If the membrane potential reaches a fixed threshold from below, the neuron will fire.

### *1.3.2.1. Leaky-Integrate-and-Fire Model*

One of most widely used threshold-fire model is the integrate-and-fire model and it has been extensively used in large spiking neural networks [17] because of the ease of implementation and the low computational cost.

The basic circuit of the integrate-and-fire model can be seen in Figure 1.9. It consists of a resistor R in parallel with a capacitor C. A pulse coming from a presynaptic neuron, passes from a low-pass RC filter before it is fed to the postsynaptic neuron.

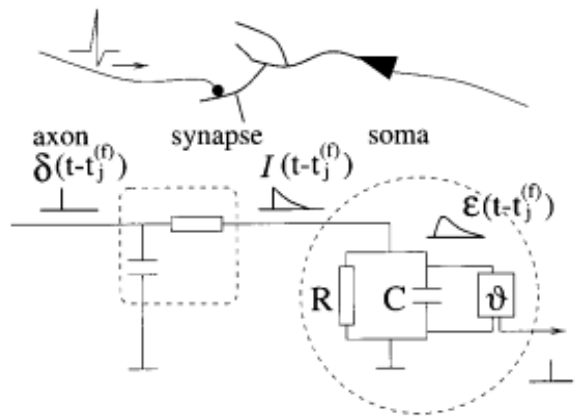

**Figure 1.9: Schematic diagram of the integrate-and-fire model [8].**

Using the Ohm's law, the schematic in the figure 1.9 can be described by the following equation:

$$I(t) = \frac{u(t)}{R} + C\frac{du}{dt} \quad (1.11)$$

And if we multiply by R and substitute with $\tau_m = RC$ we have:

$$t_m \frac{du}{dt} = -u(t) + RI(t) \quad (1.12)$$

The time constant $\tau_m = RC$ is also known as "leaky integrator" [5, 8, 18] and it represents the diffusion of ions. When a neuron receives a spike from a presynaptic neuron and its membrane does not reach the threshold, then it "leaks" back to a resting value.

Since the above equation is a first-order differential equation it cannot fully describe the spiking neuronal behavior, thus a threshold condition has to be introduced. This is expressed in equation (1.13) where the moment when the membrane potential u crosses the threshold θ from below, is described as a firing time:

$$t^{(f)} : u(t^{(f)}) = \theta \quad \text{and} \quad \left. \frac{du(t)}{dt} \right|_{t=t^{(f)}} > 0 \quad (1.13)$$

In the integrate-and-fire model the action potentials (spikes) are not described the same way as in the Hodgkin-Huxley neuron model. Here the spikes are characterized only by their firing time $t^{(f)}$, Figure 1.10.

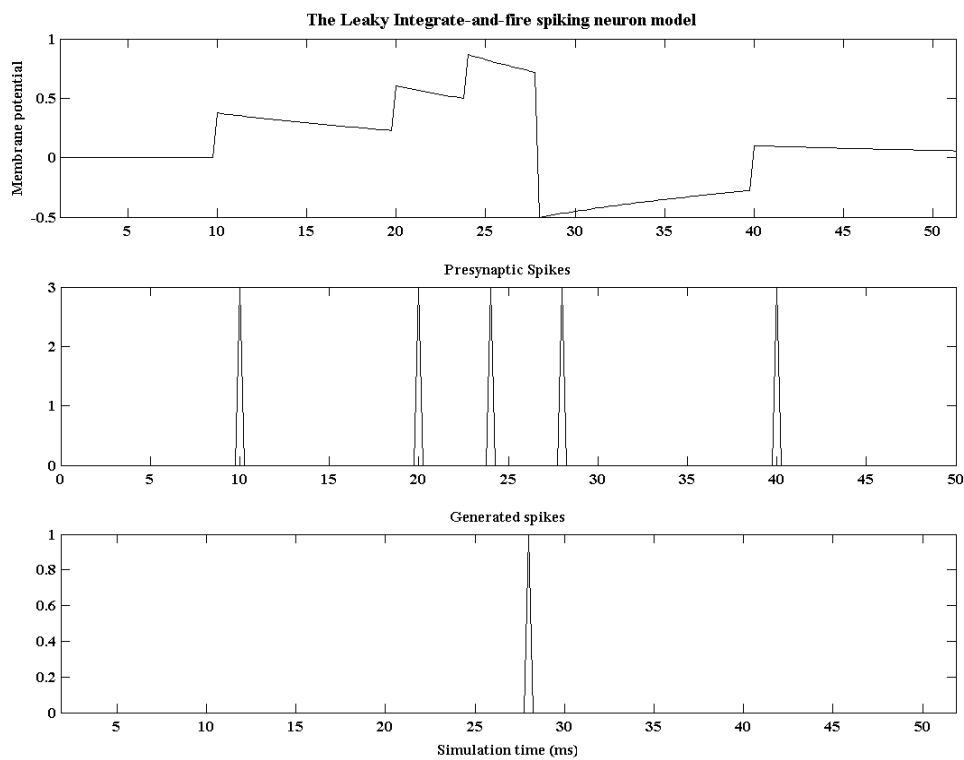

**Figure 1.10:** Simulation plot of the Leaky integrate-and-fire model using the Euler method. The threshold of the spiking neuron is set to 1. In the top graph is the membrane potential of the neuron. In the middle graph are the presynaptic spikes. Finally, the bottom plot represents the action potentials (spikes) of the postsynaptic neuron. Appendix A.3.

### *1.3.2.2. Spike Response Model – SRM*

The Spike response model (SRM) was created in order to reduce the four-dimensional Hodgkin-Huxley model into one equation. It has been proven that the SRM model can predict 90% of the Hodgkin-Huxley spike train correctly [19].

The main difference between the Spike Response Model and the leaky integrate-and-fire is that the membrane potential in the latter model is described by a differential equation and it is voltage dependent, while in the case of the SRM the membrane potential is described by *response kernels* and it is expressed at time t as an integral over the past [5].

$$u_j = \eta(t - \hat{t}_j) + \sum_i w_{ji} \sum_f \varepsilon_{ji}(t - \hat{t}_j, t - t_i^{(f)}) + \int_0^\infty k(t - \hat{t}_j, s) I^{ext}(t - s) ds \quad (1.14)$$

Where $t_i^{(f)}$ are the spikes from a presynaptic neuron i and $w_{ji}$ is the synaptic efficacy ("weight") and $I^{ext}$ is an external current, $s = t - t_i^{(f)}$ or in the case where multiple delays are used: $s = t - t_i^{(f)} - \Delta^k$ [14, 20, 21]. Finally, $\hat{t}_j$ is the last firing time of neuron j.

The neuron j fires when the membrane potential reaches the threshold value θ from below:

$$t = t_j^{(f)} \Leftrightarrow u_j(t) = \theta \text{ and } \frac{du_j(t)}{dt} > 0 \quad (1.15)$$

A dynamic threshold θ can be used instead of fixed one:

$$\theta = \theta(t - \hat{t}_j) \quad (1.16)$$

**1.3.2.2.1. The response kernels**

The response kernel $\eta(t - \hat{t}_j)$ is responsible for the after-potential of the neuron, meaning the undershoot that happens after the emission of a spike. It is characterized by the firing time $t_j^{(f)}$ when the membrane potential reaches the threshold θ.

The response kernel $\kappa(t - \hat{t}_j, s)$ is the response of the membrane potential to an input current based on the last output spike $\hat{t}_j$, Figure 1.11.

The response kernel $\varepsilon_{ji}(t-\hat{t}_j,s)$ is the time course of a postsynaptic potential after the firing of a presynaptic neuron i at time $t_i^{(f)}$, which can be either excitatory (EPSP) or inhibitory (IPSP).

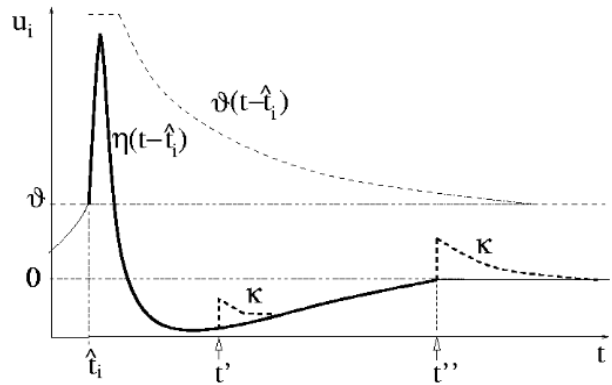

**Figure 1.11: The Spike Response Model SRM [5].**

### 1.3.2.2.2. The simplified model SRM$_0$

A very famous modification of the Spike Response Model is the simplified SRM$_0$ [5] [8], which derives from the SRM by simplifying the response kernels, Figure 1.12. In the simplified SRM$_0$ the membrane potential becomes:

$$u_j(t) = \eta(t - \hat{t}_j) + \sum_i w_{ji} \sum_{t_i^{(f)}} \varepsilon_0(t - \hat{t}_i^{(f)}) \quad (1.17)$$

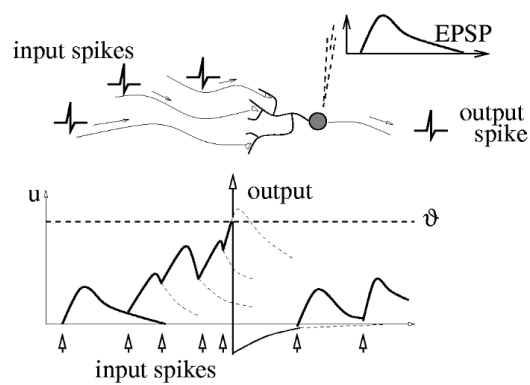

**Figure 1.12: Spike Response Model SRM$_0$ with constant threshold. Each input causes an EPSP $\varepsilon_0(s)$. When the threshold is reached a spike is emitted and the negative response kernel $\eta(s)$ is added [5].**

## 1.3.3. Comparison between the Spiking Neuron Models

The following figure shows a comparison of the computational speed and neuro-computational properties between eleven of the most widely used spiking neuron models [15].

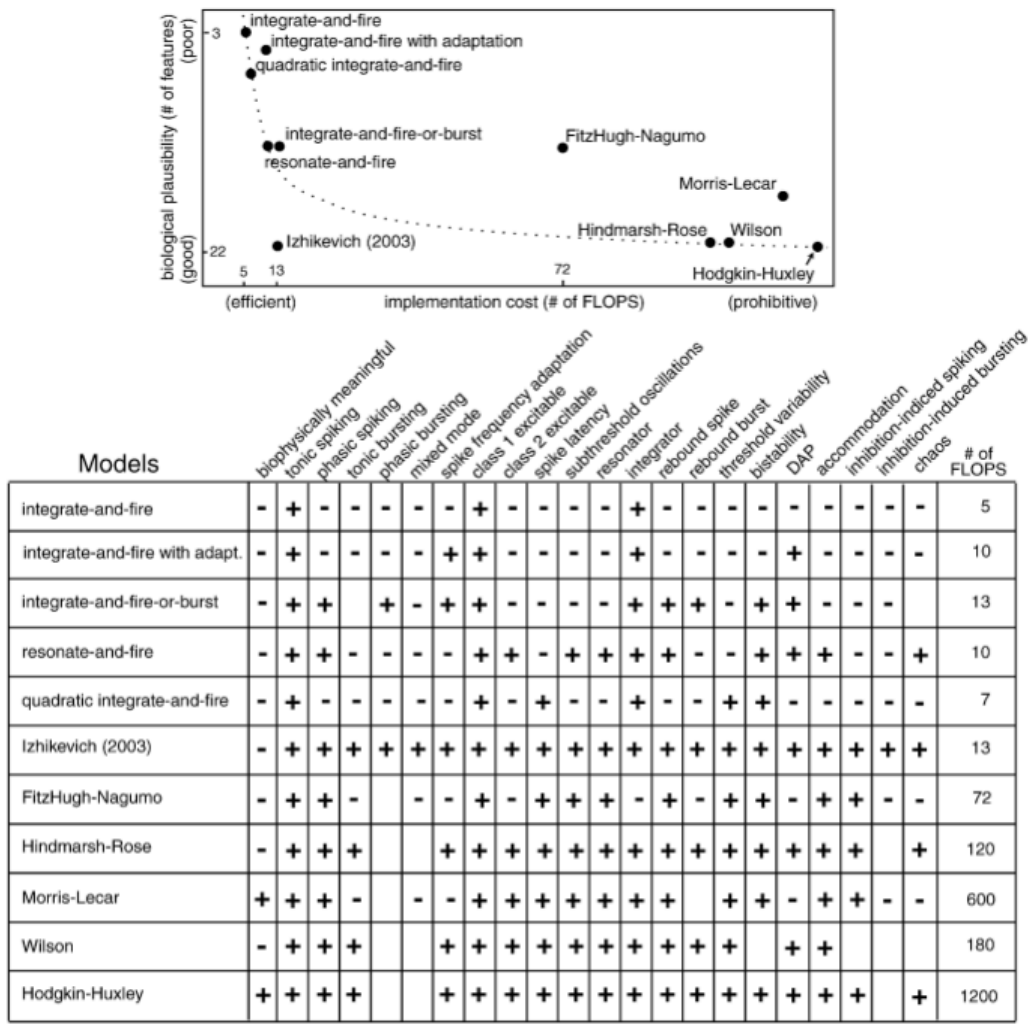

Figure 1.13: Comparison of the neuro-computational properties of spiking and bursting models. "# of FLOPS" is an approximate number of floating point operations (addition, multiplication, etc.) needed to simulate the model during a 1 ms time span. Each empty square indicates the property that the model should exhibit in principle (in theory) if the parameters are chosen appropriately, but the author failed to find the parameters within a reasonable period of time [15].

The computational cost is expressed in FLOPS (floating point operations) for 1ms of simulation time. As we can see the Hodgkin-Huxley model needs 1200 FLOPS for 1ms, which makes it unsuitable for large-scale networks. On the other hand, the integrate-and-fire model needs only 5 FLOPS for 1ms of simulation time but it can exhibit only a few of the neuro-computational. The SRM and simplified SRM$_0$ were not included in the comparison.

The Izhikevich model is able to exhibit all known neuro-computational properties and it needs 13 FLOPS for 1ms of simulation time, which makes it suitable for large neural network simulations.

**1.4. Neuronal Coding**

One of the fundamental questions of neurophysiology is how neurons encode information. A clear answer has not been found yet and all the information that is available comes from experimental results. At first, it was thought that the information was encoded as mean firing rates of a neuron. This was proven experimentally by Adrian (1926), who discovered that the firing rate of stretch receptor neurons in the muscles, was correlated to the force applied to the muscle [5]. This has been the main neuronal coding model for many years.

However, Thorpe et al. [3] proved, that humans could recognize a face in 100ms, which makes it impossible for mean firing rates coding. Furthermore, Bialek et al. [22] managed to read the visual neuron code of the fly and found out that it is formed by time-dependent signals. This proves that by using the mean firing rate all the temporal information, produced by the neurons, is lost.

Spiking neural networks can encode digital [23, 24] and analog information. The neuronal coding schemes can be divided into three categories [25]:

1. Rate coding.
2. Temporal coding.
3. Population coding

**1.4.1. Rate coding**

In rate coding the information is encoded into the mean firing rate of the neuron also known as temporal average [5]:

$$v = \frac{n_{sp}(T)}{T} \quad (1.18)$$

Where T is time window, $n_{sp}(T)$ are the number spikes emitted during the time window. There are three averaging procedures [5]: Rate as a spike count (average over time), rate as a spike density (average over several runs) and rate as a population activity (average over several neurons).

**1.4.2. Temporal coding**

In temporal coding the information is encoded in the form of spike times [26]. Hopfield [27] has proposed a method for encoding analog data into timing of the spikes with respect to an oscillatory pattern of activity. This method has been proven experimentally in the electric fish. In addition, Maass [1] proposed a method of encoding analog information in the form of firing times. A different method have been suggested by Wen and Sendhoff [28], where the input neurons encode information directly into spiking times and an additional bias neuron is used as a time reference.

**1.4.3. Population coding**

In population coding a number of input neurons (population) are involved in the analog encoding and produce different firing times. Bohte et al. [11] proposed a way of representing analog input values into spike times using population coding. Multiple Gaussian Receptive Fields (GRF) were used so that the input neurons will encode an input value into spike times, Figure 1.14.

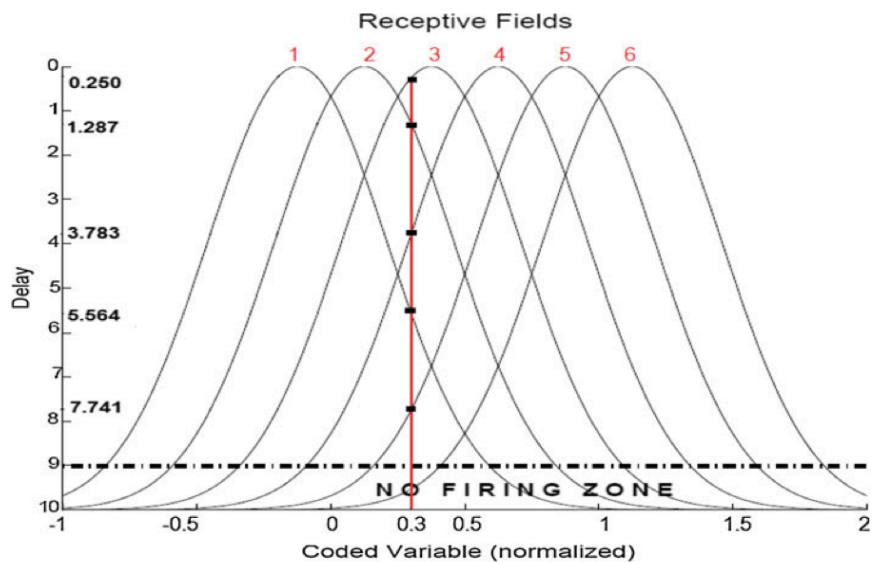

**Figure 1.14: Encoding with Gaussian Receptive Fields. The horizontal axis represents the real input data, the vertical axis represent the firing times of the input neurons to an input value 0.3 [25].**

Firstly the range of the input data has to be calculated. Then the values $I_{max}$ and $I_{min}$, which are the maximum and minimum values of the input data, have to be defined. Furthermore, the number of GRF neurons that are going to be used has to be chosen through the m variable. Lastly, the center of each GRF neuron is calculated from $C_i$ while the width of each GRF neuron is calculated by $\sigma_i$ [25]:

$$C_i = I_{min} + (\frac{2i-3}{2})(\frac{I_{max} - I_{min}}{m-2}) \quad (1.19)$$

$$\sigma_i = \frac{1}{\gamma} \frac{I_{max} - I_{min}}{m-2} \quad (1.20)$$

Where γ is constant number usually around 1.5. A threshold value has to be used so that GRF neurons, that are below the threshold, should not fire. In the example of Figure 1.14 the analog value 0.3 is encoded into firing times of neuron 3 (0.250ms), neuron 2 (1.287ms), neuron 4 (3.783ms), neuron 1 (5.564ms) and neuron 5 (7.741ms). Neuron 6 does not emit a spike because it's below the threshold.

**1.5. Learning methods**

The weights $w^k_{ji}$ between a presynaptic neuron i and a postsynaptic neuron j do not have fixed values. It has been proved through experiments that they change, thus affecting the amplitude of the generated spike. If the synaptic strength is increased it is called long term potentiation (LTP) and if the strength decreases it is called long term depression (LTD). The procedure of the weight update is called learning process and it can be divided into two categories: supervised and unsupervised learning.

**1.5.1. Unsupervised Learning**

In 1949 Hebb formulated the famous Hebb law: "When an axon of cell A is near enough to excite cell B or repeatedly or persistently takes part in firing it, some growth process or metabolic change takes place in one or both cells such that A's efficiency, as one of the cells firing B, is increased".

Hebb's law is modified so that the weights are updated based on the pre and postsynaptic activity of the neurons.

### 1.5.1.1. Hebbian model – Spike Time Dependent Synaptic Plasticity (STDP)

In the following figure we can see the experimental results of Bi and Poo [5] on the hippocampal neurons.

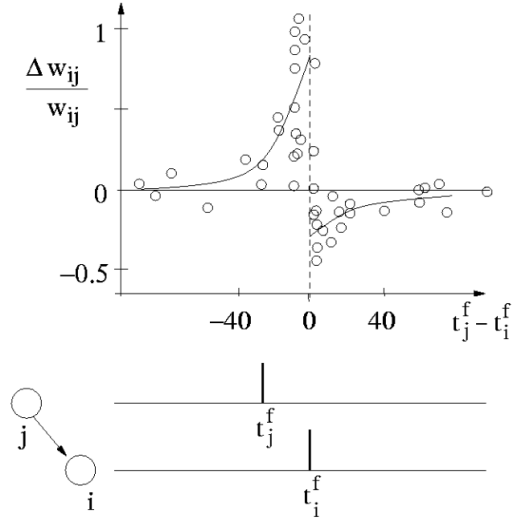

**Figure 1.15: The weights are changing only if the firing times of neurons j and i are close to each other. Data taken from the experiments of Bi and Poo (1998) [5].**

Where neuron j is the presynaptic neuron, neuron i is the postsynaptic neuron and $t_j^f$ is the presynaptic fire time and $t_i^f$ is the postsynaptic fire time.

Furthermore, Bi and Poo [5] found out that the synaptic efficacy $\Delta w_{ij}$ is a function of the spike times of the presynaptic and postsynaptic neurons. This is called Spike Timing-Dependent Plasticity (STDP) [29].

A way to calculate the synaptic weight updates has been proposed by Gerstner et al. [5] with the use of exponential learning windows:

$$\Delta w = \begin{cases} A_+ \exp(s/\tau_1) \; for \; s < 0 \\ A_- \exp(s/\tau_2) \; for \; s > 0 \end{cases} \quad (1.21)$$

Where $s = t_j^{(f)} - t_i^{(f)}$ is the time difference between presynaptic and postsynaptic firing times. The $\tau_1$ and $\tau_2$ are constants and the $A_+$ and $A_-$ are used for stability issues in order to cap the weights to a maximum and minimum value, Figure 1.16.

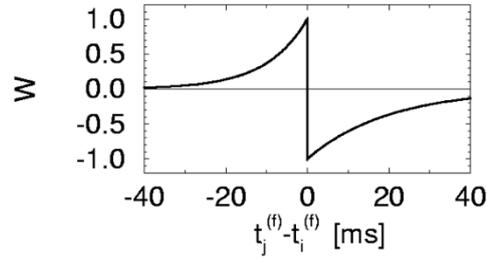

**Figure 1.16: The exponential learning window as a function the difference between the presynaptic and the postsynaptic firing times. A+=1, A-=-1, τ1=10ms, τ2=20ms [5].**

Numerous methods have been proposed in order to overcome the need of capping the weights to maximum and minimum values for unsupervised learning.

One of these methods is the Synaptic Weight Association Training (SWAT) [30]. In SWAT method the STDP is combined with the Bienenstock-Cooper-Munro (BCM) method, where a sliding threshold is used for weight stability. This model uses a feed-forward topology similar to the hippocampus where an inhibitory and excitatory synapse is between every presynaptic and postsynaptic neuron.

### *1.5.1.2. Local Hebbian delay-learning*

The local Hebbian delay-learning method uses a winner-take-all rule [31] implemented where multiple sub-synapses with multiple delays (delayed spiking neurons) are used, as in Figure 1.4.

This learning method is very popular in the cases of unsupervised clustering tasks [11, 25, 32] and since it is a competitive learning mode only the weights of the winner neurons are updated [25]. The learning rule that is used, is a Gaussian function L(Δt), that takes the difference between the presynaptic and postsynaptic neuron firing times as input, in order to update the weights, equations 1.22-1.24. By using this method in a clustering task, each output neuron becomes a peusdo-RBF centre [25, 11].

$$\Delta w_{ij}^{k} = \eta L(\Delta t_{ij}) \quad (1.22)$$

$$L(\Delta t) = (1+b)\exp[\frac{(\Delta t - c)^2}{2(k-1)}] - b \quad (1.23)$$

$$k = 1 - \frac{v^2}{2\ln[\frac{b}{1+b}]} \quad (1.24)$$

Where $\Delta w_{ij}$ represent the amount of change of the weights, $L(\Delta t_{ij})$ is the learning function, η is the learning rate, ν is the width of the learning window, Δt is the difference between the presynaptic j and postsynaptic neuron i firing times, b is for the negative update of a neuron and c represents the peak of the learning function [25], Figure 1.17.

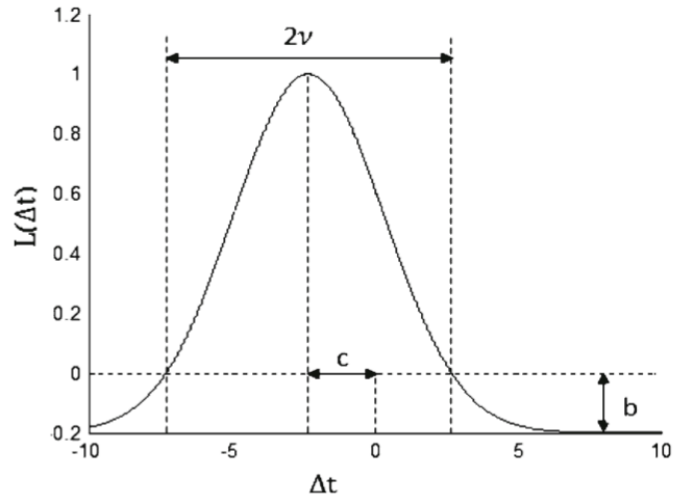

Figure 1.17: The learning rule L(Δt) versus Δt with b=0.2, c=-2.3 and ν=-5 [25].

Similar to the STDP described in the previous section, the weights are capped to a minimum and maximum value, 0 and $w_{max}$ [11, 25].

**1.5.2. Supervised Learning**

One of the reasons that lead to a tremendous increase in research of the sigmoidal neural networks, was the error Backpropagation training algorithm. The Backpropagation algorithm is a supervised learning mode that is based on the gradient descent method to minimize the output error.

As described in the previous sections, spiking neural networks are more biologically close to the real neural networks and more powerful than the previous generations of artificial neural networks. However, their computational power cannot be fully used due to the limited number of supervised learning methods. The main reasons for this, is that spiking neural networks are discontinuous in contrast to their continuous predecessors. There are numerous methods proposed for supervised learning.

### *1.5.2.1. SpikeProp*

Bohte et al. [14] proposed a supervised learning algorithm based on the Backpropagation method of the sigmoidal artificial neural networks. This method is called Spikeprop and is designed for spiking neural networks with multiple delayed synapses, Figure 1.4.

The spiking neural network architecture is a feed-forward network that can have one or more hidden layers. For this case the network is similar to Figure 1.4 with one hidden layer. H is the input layer, I is the hidden layer and J is the output layer.

Bohte et al. [14] used a spike response model (SRM) to describe the membrane potential of a spiking neuron without a refractoriness response kernel, since the neurons are allowed to fire only once:

$$u_j(t) = \sum_{i \varepsilon \Gamma_j} \sum_{k=1}^{m} w_{ij}^k \varepsilon(t - t_i - d^k) \quad (1.25)$$

The purpose of the algorithm is to train the network to a desired output firing times for specific input patterns, by minimizing the error. The error is described as:

$$E = \frac{1}{2} \sum_{j \varepsilon J} (t_j^a - t_j^d)^2 \quad (1.26)$$

Where $t^a_j$ is the actual firing time and $t^d_j$ is the desired firing time.

- The weight update for the neurons of the <u>output layer</u> is

$$\Delta w_{ij}^k = -\eta y_i^k(t_j^a) \delta_j \quad (1.27)$$

Where η is the learning rate, and $\delta_j$ is

$$\delta_j = \frac{(t_j^d - t_j^a)}{\sum_{i\varepsilon\Gamma_j}\sum_l w_{ij}^l(\partial y_i^l(t_j^a)/\partial t_i^a)} \quad (1.28)$$

- The weight update for the neurons of the <u>hidden layer</u> is

$$\Delta w_{hi}^k = -\eta y_h^k(t_i^a)\delta_i = -\eta \frac{y_h^k(t_i^a)\sum_j\{\delta_j \sum_k w_{ij}^k(\partial y_i^k(t_j^a)/\partial t_i^a)\}}{\sum_{n\varepsilon\Gamma_i}\sum_l w_{ni}^l(\partial y_n^l(t_i^a)/\partial t_i^a)} \quad (1.29)$$

where $\delta_i$ is:

$$\delta_i = \frac{\sum_{j\varepsilon\Gamma^i}\delta\{\sum_k w_{ij}^k(\partial y_i^k(t_j^a)/\partial t_i^a)}{\sum_{h\varepsilon\Gamma_i}\sum_l w_{hi}^l(\partial y_h^l(t_i^a)/\partial t_i^a)} \quad (1.30)$$

The mathematical proofs of this algorithm can be found in the reference [14]. Finally, the Spikeprop algorithm can be summarized in the following table:

**Table 1.2: The Spikeprop algorithm [14].**

*SpikeProp algorithm*

Calculate $\delta_j$ for all outputs according to (12)

For each subsequent layer $I = J - 1 \ldots 2$
  Calculate $\delta_i$ for all neurons in $I$ according to (17)

For output layer $J$, adapt $w_{ij}^k$ by $\Delta w_{ij}^k = -\eta y_i^k(t_j)\delta_j$ (14)
For each subsequent layer $I = J - 1 \ldots 2$
  Adapt $w_{hi}^k$ by $\Delta w_{ij}^k = -\eta y_h^k(t_i)\delta_i$ (18)

One of the disadvantages of the Spikeprop algorithm is that it takes a big number of iterations until it converges. A number of techniques have been proposed to speedup its process; some of them are the RProp and QuickProp algorithms [21].

In Table 1.3 and 1.4 a comparison of the number of iterations between the three training algorithms can be seen; for traditional neural networks (ANN) and for spiking neural networks (SNN), and for two different datasets. One thing that is noticeable is that the SNN converges with much less iterations than the traditional sigmoidal artificial neural networks (ANN).

Table 1.3: The XOR dataset. Number of iterations to 0.5 Mean Square Error [21].

|  | ANN | SNN |
|---|---|---|
| **Spikeprop** | 2750 | 127 |
| **RProp** | 386 | 29 |
| **QuickProp** | 51 | 31 |

Table 1.4: The Fisher dataset. Number of iterations to 0.5 Mean Square Error [21].

|  | ANN | SNN |
|---|---|---|
| **Spikeprop** | 1370 | 222 |
| **RProp** | 76 | 25 |
| **QuickProp** | 79 | 53 |

To conclude, Spikeprop uses the Backpropagation method, which is based on the gradient decent so a number of assumptions have been taken into consideration in order to overcome the discontinuity of spiking neural networks.

One of these assumptions is that the neurons are allowed to fire once [30, 28, 21], thus this method cannot take the full advantage of temporal processing. Furthermore, this training algorithm works only for small learning rates (η) and its performance depends on the initial parameters [28]. Finally, since neurons are allowed to fire only once, only the time-to-first spike coding scheme can be used [30].

*1.5.2.2. Spikeprop for multiple spikes*

In order to overcome the disadvantage of Spikeprop algorithm, Booij et al. [20] proposed a modification of the Spikeprop algorithm so that each neuron could fire multiple times. Similar to the Spikeprop, the weights are updated to achieve the desired output firing times, by minimizing the error. A feed-forward network was used with multiple delays and multiple synapses per connection, as seen in Figure 1.4.

In this method the spike response model (SRM) was used as in the original Spikeprop [14], however, the main difference is the membrane potential. In this case is described as:

$$u_j(t) = \sum_{t_j^f \varepsilon F_j} \eta(t - t_j^f) + \sum_{i \varepsilon \Gamma_j} \sum_{t_i^f \varepsilon F_i} \sum_k w_{ji}^k \varepsilon(t - t_i^f - d_{ji}^k) \quad (1.31)$$

Where $w^k_{ji}$ is the weight of synapse k from a presynaptic neuron i to a postsynaptic neuron j. Equation 1.31 has a mistake that is biologically incorrect: each neuron's refractoriness kernel η, is the sum of all the neurons in the same layer[41]. Finally, the error is expressed as in the original Spikeprop algorithm, equation 1.26.

- In Spikeprop the backpropagation rule was:

$$\Delta w_{ih}^k = -k \frac{\partial E}{\partial w_{ih}^k} \quad (1.32)$$

However, since a neuron can fire multiple times, Booij et al.[20] modified the above equation to:

$$\Delta w_{ih}^k = -k \sum_{t_i^f \varepsilon F_i} \frac{\partial E}{\partial t_i^f} \frac{\partial t_i^f}{\partial w_{ih}^k} \quad (1.33)$$

In addition, the authors added two special rules to overcome the discontinuity of the threshold function. The first rule was that they used a lower bound on the gradient of the potential to avoid very big changes in the weights and instability. The second rule was that if a postsynaptic neuron did not fire due to very small weight, the weight had to be increased by a small value.

This algorithm was tested on the XOR benchmark and the Poisson spike trains classification problem with the same network architecture as in Spikeprop [14] and proved able to solve both of them.

One of the computational capabilities of spiking neural networks using Spikeprop with multiple spikes was that the authors were able to solve the XOR benchmark without a hidden layer. This is impossible with the sigmoidal artificial neural networks, since networks with one layer can only solve linear separable problems. However, some additional special rules were taken into consideration. For example, the learning rate κ was set to a very small value and the convergence time proved too big ($10^6$ cycles). Finally, as the authors pointed out, this method would not be robust in the presence of noise.

## 1.6. The proposed supervised training algorithm

### 1.6.1. Discussion on the existing supervised learning methods

Even though unsupervised learning methods have been widely used in spiking neural networks [23-25, 11] and produced very good results, things are not so great when it comes to supervised learning. In fact, there is still ongoing research on this field.

Kasinski et al. [33] reviewed some of the supervised learning methods of spiking neural networks. The summary of his review can be seen in Table 1.5. As one can see, only four of the eight methods described are allowed to train neurons that can fire multiple spikes.

A very interesting approach to the supervised learning would be the use of evolutionary algorithms (EA), since they can handle the discontinuity of spiking neurons in contrast to the gradient decent method. In most cases, the evolutionary algorithms that were used was the evolutionary strategies (ES) method [33]. This method does not need to encode the information to binary; real numbers can be used instead.

In reference [28], a method using genetic algorithms (GA) for supervised training was proposed. This method was able to train both synaptic weights and delays in a multi sub-synapses, multi delay network, as in Figure 1.4. However, the spiking neurons were allowed to fire only once, thus only time-to-first spike coding scheme could be used.

Table 1.5: A list of the supervised learning techniques that were reviewed [33].

| No | Author/Method | Approach | Presented Tasks | Coding Scheme | References |
|---|---|---|---|---|---|
| 1 | S. Bohte/*SpikeProp* | Gradient estimation | Classification | Time-to-first-spike | (Bohte, 2003; Bohte *et al.*, 2000; 2002; Moore, 2002; Schrauwen and Van Campenhout, 2004; Xin and Embrechts, 2001) |
| 2 | J. Pfister, D. Barber, W. Gerstner | Statistical approach | Spike sequence learning, Classification | Precise spike timing | (Barber, 2003; Pfister *et al.*, 2003; 2005) |
| 3 | A. Carnell, D. Richardson | Linear algebra formalisms | Spike sequence learning | Precise spike timing | (Carnell and Richardson, 2004) |
| 4 | A. Belatreche, L. Maguire, T. McGinnity, Q. Wu | Evolutionary strategy | Classification | Time-to-first-spike | (Belatreche *et al.*, 2003) |
| 5 | J. Sougne | Learning in synfire chains | Single temporal interval learning | Relative spike time | (Sougne, 2001) |
| 6 | B. Ruf, M. Schmitt | Supervised Hebbian learning | Classification | Time-to-first-spike | (Ruf, 1998; Ruf and Schmitt, 1997) |
| 7 | R. Legenstein, Ch. Naeger, W. Maass | Supervised Hebbian learning | Spike sequence learning, Input-output mapping | Precise spike timing | (Legenstein *et al.*, 2005) |
| 8 | F. Ponulak/*ReSuMe* | Remote supervision | Spike sequence learning, Input-output mapping, Neuron model independence, Real-life applications | Precise spike timing | (Kasiński and Ponulak, 2005; Ponulak, 2005; Ponulak and Kasiński, 2005) |

**1.6.2. Discussion on limited synaptic precision and the proposed training algorithm**

Spiking neural networks are more suitable for parallel processing compare to traditional neural networks, due to the fact that they are asynchronous. The majority of hardware implementations of spiking neural networks have been done on Field Programmable Gate Arrays (FPGA) because they offer scalability, low cost, and fast processing times.

However, when it comes to hardware implementation of neural networks, limited weight precision is an important factor that has to be taken into consideration. The weights can only be represented by a finite number of bits and if a neural network can be trained with limited precision, this would result in reduction of size, complexity and cost [34-36].

Different limited precision schemes have been used for the traditional artificial neural networks, from integer weights [35] to lookup tables for the sigmoid activation function [34]. In the latter case the quantize backpropagation step-by-step method [34] has shown a 70% speedup compared to the conventional neural networks. However, limited precision has not been applied to spiking neural networks yet. It would be very interesting to investigate spiking neural networks with limited synaptic precision as the results would be helpful for future hardware implementations.

For these reasons a supervised training algorithm for limited precision feed-forward spiking neural networks will be developed and it will be able to train both synaptic weights and delays and at the same time allow each neuron to emit multiple spikes.

# 2. Matlab simulation for fully connected feed-forward Spiking Neural Networks

## 2.1. Introduction

As there is no Matlab toolbox for modeling the Spiking Neural Networks, a program for simulating fully connected feed-forward spiking neural network was developed. The program was able to design neural networks with single synapse per neuron, Figure 2.1. It was used to develop the proposed supervised training algorithm based on Evolutionary Algorithms, where the weights and delays of the synapses were trained using limited precision to achieve the desired output spiking times.

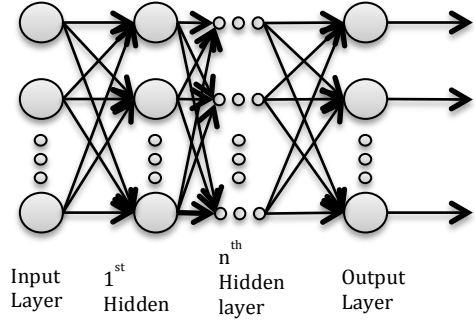

**Figure 2.1: A fully connected feed forward architecture.**

Each spiking neuron's membrane potential was described using the spike response model $SRM_0$ as follows:

$$u_j(t) = \rho(t - t_j^{(F)}) + \sum_{i=1}^{N_{l+1}} \sum_{g=1}^{G_i} w_{ji} \varepsilon(t - t_i^{(g)} - d_{ji}) \quad (2.1)$$

$$\varepsilon(t_e) = \frac{t_e}{\tau} e^{1-\frac{t_e}{\tau}} H(t_e) \quad (2.2)$$

$$\rho(t_p) = -4\vartheta e^{-\frac{t_p}{\tau_R}} H(t_p) \quad (2.3)$$

$$t = t_j^{(F)} \Leftrightarrow u_j(t) = \vartheta \quad and \quad \frac{du_i(t)}{dt} > 0 \quad (2.4)$$

Where $G_i$ are the total spikes from the previous layer fired at time $t_i^{(g)}$. The $w_{ji}$ and $d_{ji}$ represent the weights and delay time of each synapse. Function $\varepsilon(.)$ is the unweight internal response of the postsynaptic neuron to a single spike. Function $\rho(.)$ is the refractoriness function and $t_j^{(F)}$ is the time of the most recent spike. The function H(.) is a Heaviside function. Lastly, equation (2.4) defines the threshold conditions of emitting a spike.

This model is based on [41]; however, it is modified for neural networks with single synapse per neuron. Furthermore, this model is able to handle multiple spikes from all ($N_{l+1}$) presynaptic neurons to the $j^{th}$ neuron in layer l, thus taking full advantage of the spatial-temporal coding power.

Finally, the spiking neural network architecture is chosen by an array variable named Topology. The size of this array represents the number of layers and its values represent the number of neurons in each layer. For example, Topology=[2 2 1] means that there are 3 layers: 1 input layer with 2 neurons, 1 hidden layer with 2 neurons and 1 output layer with 1 neuron, Figure 2.2.

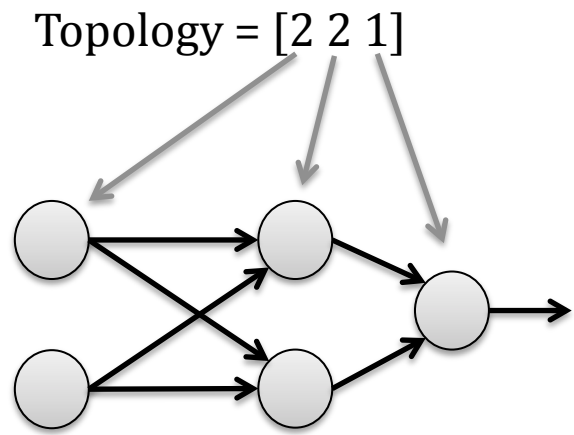

**Figure 2.2: The Topology array defines the network architecture.**

The hierrarchy chart of the programme can be seen in Figure 2.3, while the specification analysis, pseudo-English codes and data tables can be observed in Appendix K.

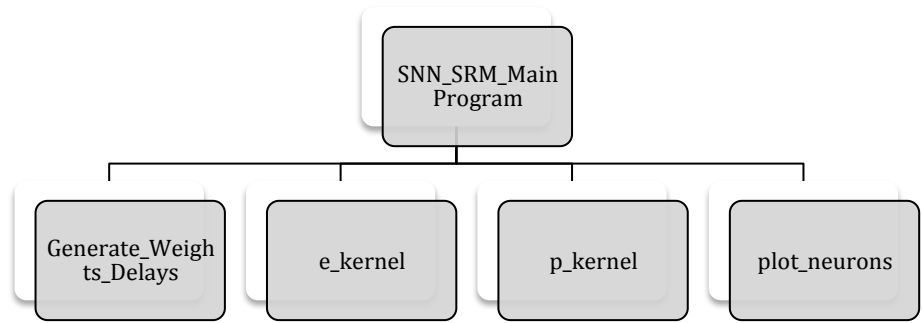

**Figure 2.3: Hierarchy chart of the Matlab spiking neural network program.**

## 2.2. Validation of the Matlab programme

In this section the Matlab program is tested on two different topologies, section 2.2.1 and 2.2.2. The parameters for each simulation can be seen in Table 2.1 and Table 2.3 while the input spikes are shown in Table 2.2 and Table 2.4. Finally, each neuron's membrane potential can be seen in the following plots.

### 2.2.1. Network architecture, Topology=[2 1]

The network architecture that was tested can be seen in Figure 2.4, while the membrane potential of the output neuron can be observed in Figure 2.5.

Table 2.1: Simulation settings.

| Simulation time | 100 ms |
|---|---|
| Topology | [2 1] |
| Time step | 0.01 |
| Tau | 3 |
| tauR | 20 |
| Threshold | 1.5 |

Table 2.2: 3 Input spikes from neurons N1 and N2 in ms.

| N1 | 0 | 10 | 0 |
|---|---|---|---|
| N2 | 15 | 60 | 70 |

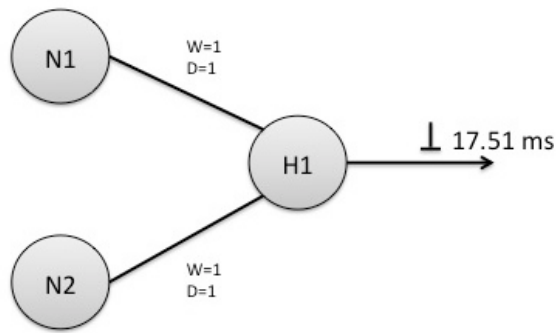

Figure 2.4: Network architecture Topology=[2 1]. W and D are the weights and delays of the synapses. One output spike is emitted at 17.51 ms.

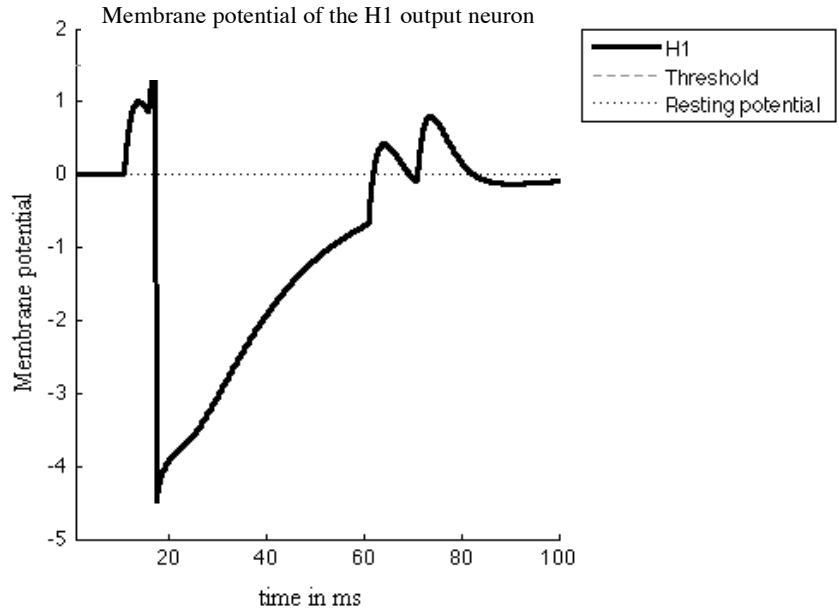

**Figure 2.5: Membrane potential of the H1 neuron.**

### 3.2.2. Network architecture, Topology=[2 2 1 3]

This time, the network architecture that was tested can be seen in Figure 2.6, while the membrane potential of the output neuron can be observed in Figure 2.7.

**Table 2.3: Simulation settings.**

| Simulation time | 100 |
|---|---|
| Topology | [2 2 1 3] |
| Time step | 0.01 |
| tau | 3 |
| tauR | 20 |
| Threshold | 1.5 |

**Table 2.4: 3 Input spikes from neurons N1 and N2.**

| N1 | 0 | 10 | 0 |
|---|---|---|---|
| N2 | 15 | 60 | 70 |

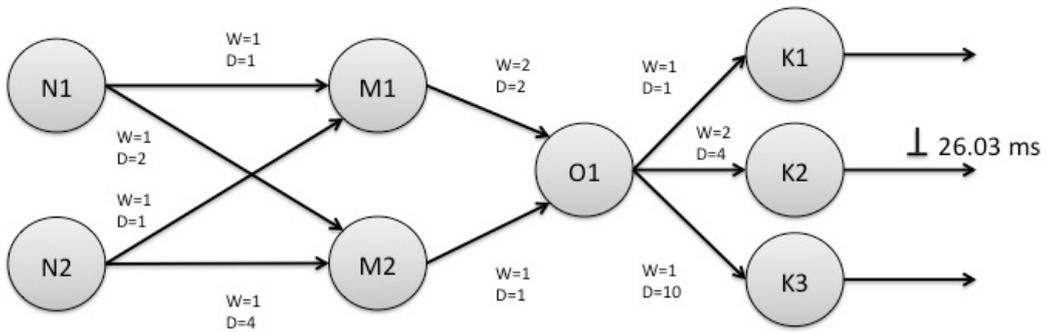

**Figure 2.6: Network architecture = [2 2 1 3]. W and D are the weights and delays of the synapses. One output spike is emitted at 26.03ms from neuron K2.**

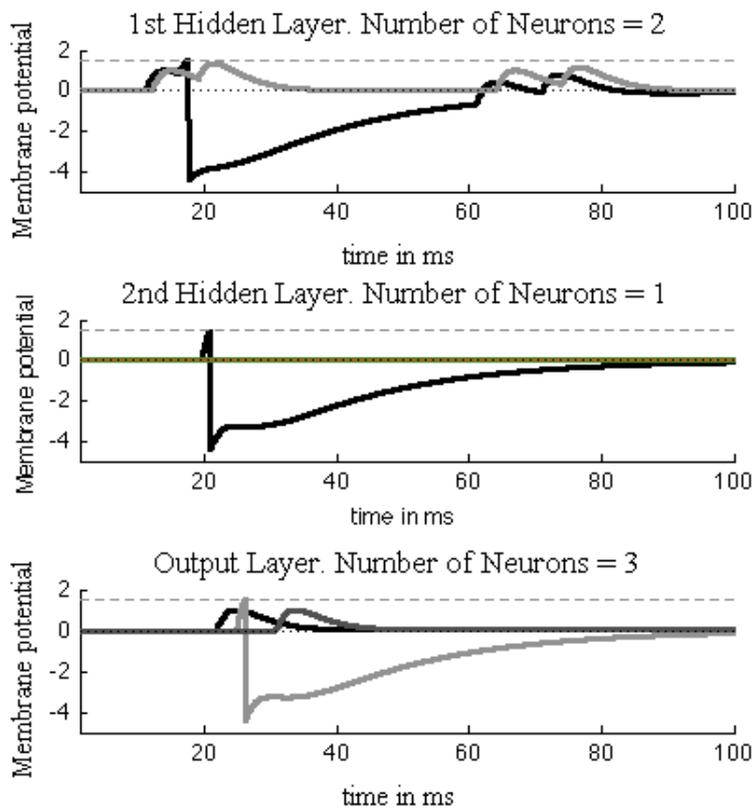

**Figure 2.7: The membrane potential of all neurons. The 1st graph is the 1st hidden layer, the 2nd graph is the second hidden layer and the 3rd is the output layer.**

# 3. A genetic algorithm for supervised learning

## 3.1. Introduction

Genetic algorithms (GA) is a search and optimization method that is based on natural selection. In general, a genetic algorithm is characterized as "any population-based model that uses selection and recombination operators to generate new sample points in a search space" [37].

Furthermore, in GA each solution to the problem is encoded as a chromosome. Different encoding schemes exist; however, the binary one is the most commonly used. After the encoding, a random population of chromosomes is generated, where its size is set to be a multiple of the dimensionality of the problem [38]. Subsequently, the fitness function of each chromosome is calculated based on their objective function and the selected fitness transformation scheme. The chromosomes are, then, selected for mating based on their fitness functions. After that, their genetic material will be mixed based on a crossover probability in order to produce offspring (new solutions). Mutation is then applied to the offspring with a probability $p_m$, also known as mutation rate and by using this, a bit of the generated offspring, is flipped. The mutation operator is very important because it offers new regions to the solution space and prevents from premature convergence [38]. Then the population is updated based on the generated offspring and the best individuals from the previous population, which pass unconditionally to the next population by using the elitism operator [38]. Finally, since the Genetic Algorithms is a stochastic algorithm, a termination criterion has to be established.

Some of the advantages of the Genetic algorithms are [38, 39]:

- They use parameter encoding in order to operate and not the parameters themselves.
- They use a set of points simultaneously in their search, instead of single points which makes them more suitable for parallel computing.
- They can handle efficiently and robustly high dimensional search spaces with discontinuous, multimodal and noisy objective functions.

- They present a set of solutions instead of a single solution.
- They do not require derivative information and are able to avoid local minima.

However, some of the disadvantages are [38]:

- They are time consuming especially when the population is too large or when too many objective function evaluations are required.
- Since they are based on stochastic algorithms a convergence is not always guarantee.
- They are greatly affected by their initial parameters.

## 3.2. The proposed genetic algorithm

As described in the first chapter of this thesis, one of the major drawbacks of the spiking neural networks is the lack of a good supervised learning algorithm. That is because the Backpropagation algorithm of the traditional artificial neural networks cannot be directly implemented, since it is a gradient-based method and the spike response model is discontinuous (threshold function), equation (3.4).

The proposed supervised training algorithm is designed for single synapse per neuron and it is based on a genetic algorithm that trains both weights and delay times of the synapses in order to achieve the desired output spike-times. Furthermore, it can accept multiple spikes from all neurons. The flow chart of the proposed genetic algorithm can be seen in Figure 3.1 and it is followed by a description of each block.

Before any training takes place the weights and delay times of the synapses have to be encoded into genotypes. Two different coding schemes were used in this thesis: The first one was the 3 bit integer values for the delay times and 3 bit with one binary decimal place for the weight values, while the second one was the 3 bit integer values for the delay times and 3 bit integer values for the weights. Both methods are described in the following sections.

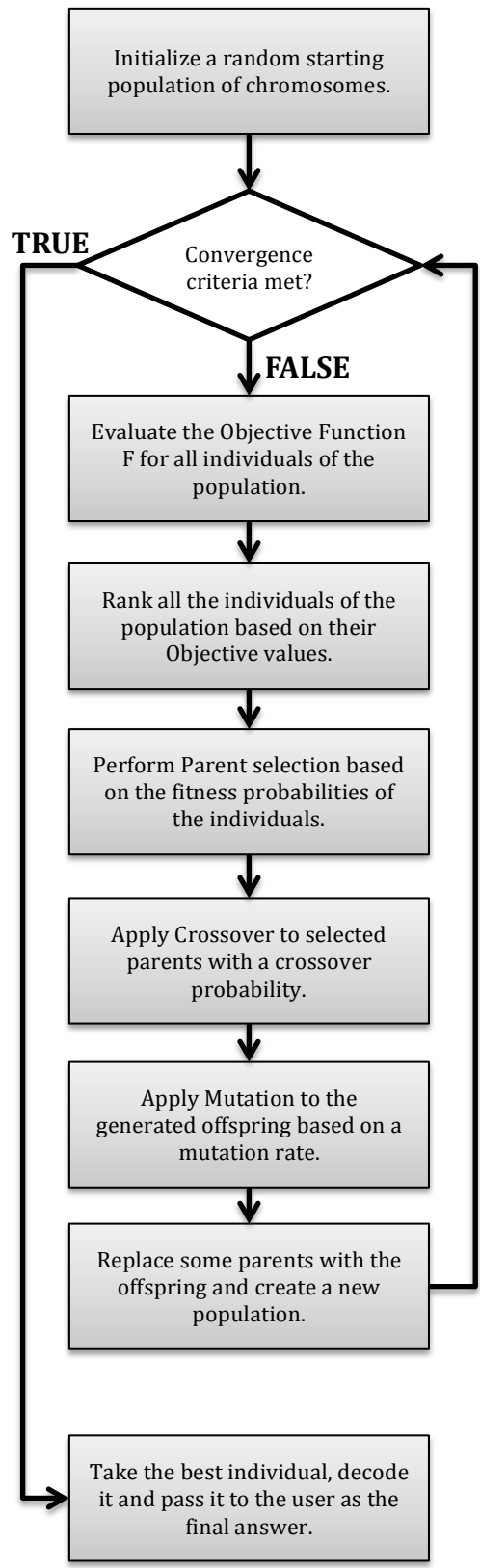

**Figure 3.1: The flow chart of the proposed genetic algorithm**

### 3.2.1. The objective function

The objective function that was used for this algorithm was the least mean-squared error (LMS) function but the mean-squared error (MSE) function was used as well:

$$LMS = \frac{1}{2} \sum_{m=1}^{InputPatterns} (t_j - t_j^d)^2 \quad (3.1)$$

$$MSE = \frac{1}{InputPatterns} \sum_{m=1}^{InputPatterns} (t_j - t_j^d)^2 \quad (3.2)$$

Where $t_j$ is the actual output spike-time and $t_j^d$ is the desired spike-time. The Matlab program from chapter 2 was used; where for a given input spike-pattern it returned an output spike.

### 3.2.2. Ranking of the individuals

All individuals were ranked based on their objective values with the first one being the best individual. For the ranking process the Baker's ranking scheme was used [38]:

$$p_i = \frac{1}{N}[\eta_{max} - (\eta_{max} - \eta_{min})\frac{i-1}{N-1}] \quad (3.3)$$

Where i is the ranked index of the individuals, $\eta_{min}=2-\eta_{max}$ and $\eta_{max}\in[1,2]$ and $\sum p_i=1$.

One of the advantages of using the ranking method is that the selective pressure can be directly controlled ($\eta_{max}$) without the need of scaling, Figure 3.2.

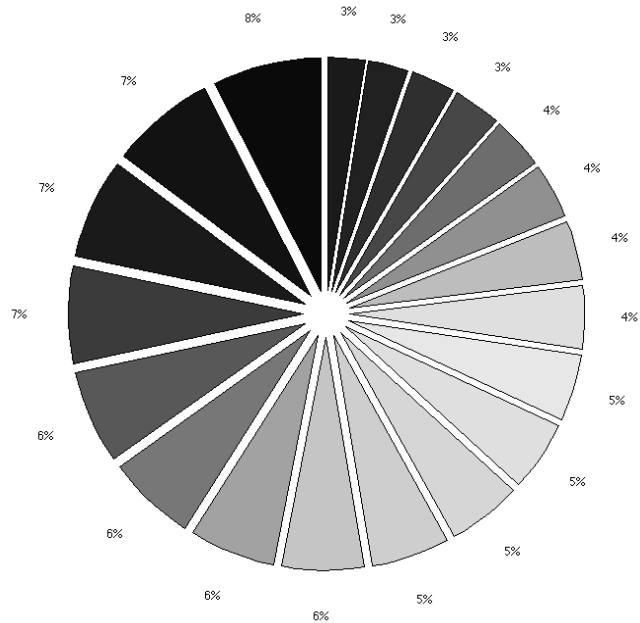

**Figure 3.2: The fitness probabilities of a population of 20 individuals. The individuals are ranked with the first one being the best one. The true values of the fitness probabilities are the ones seen divided by 100 since $\sum p_i = 1$.**

**3.2.3 Parent selection scheme**

The next step to ranking the individuals was the parent selection step where the individuals were chosen to mate and create offspring (new solutions) by swapping their genes. For this algorithm the Stochastic Universal Sampling scheme was used because it produced better results than the Roulette-selection and it also provided zero bias and minimum spread.

This method is similar to the Roulette-selection, however, here a number of markers are placed. The total number of markers is equal to the individuals that are going to be selected and they are equally spaced to each other. Further information on selection schemes can be found in [42].

**3.2.4. The Crossover operator**

The new offspring (new solutions) were generated with the crossover operator. For this algorithm the Uniform Crossover method was chosen where two individuals exchange genetic material based on a randomly generated mask, Figure 3.3.

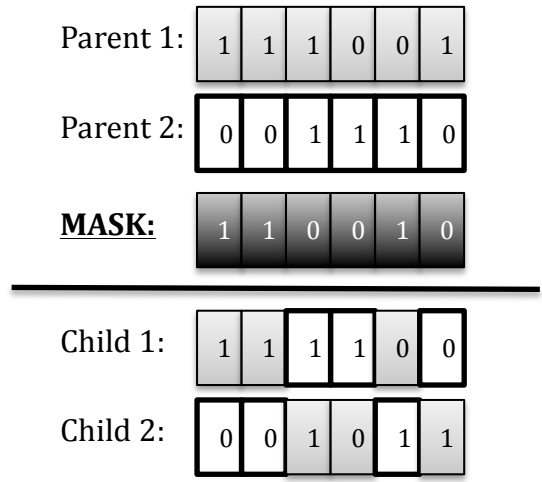

**Figure 3.3: Uniform Crossover.**

### 3.2.5. The Mutation operator

The standard Bit-flip mutation was used where each bit of the generated offspring can change from 0 to 1 or from 1 to 0 based on a mutation rate probability.

The mutation operator offers exploration of the solution space [38].

### 3.2.6. The Population model

The proposed algorithm used a generational population with an elitism operator. In the generational population or non-overlapping population, a new N size population is created and replaces the old one. The elitism operator makes sure that the best individuals will pass to the next generation unconditionally.

### 3.2.7. The convergence criteria

This algorithm used the same convergence criteria as [43], which was the mean-squared error (MSE) of 0.25, equation 3.2.

### 3.3. The encoding/decoding schemes

As stated before, two encoding/decoding schemes are investigated in this project:

1. The 3 bit with one binary decimal place weights.
2. The 3 bit integer weights.

In both of these schemes the synaptic delay values were integers. Moreover, a trade-off between complexity and biological applicability had to be made. Biological neural networks use excitatory and inhibitory neurons and the synaptic weights have positive values. However, this algorithm uses only excitatory neurons and positive and negative values for the synaptic weights like traditional artificial neural networks do. Both coding schemes are discussed in the following sections.

### 3.3.1. The 3 bit with one binary decimal place encoding/decoding scheme

The encoding/decoding of the synaptic delay times can be seen in Table 3.1, while the encoding/decoding scheme of the weights can be observed in Table 3.2. In both cases, genotype is the encoded form of the solution (phenotype).

**Table 3.1: Encoding/Decoding of the synapse delay times in ms.**

| Genotype | Phenotype |
|----------|-----------|
| 000 | 1 |
| 001 | 2 |
| 010 | 3 |
| 011 | 4 |
| 100 | 5 |
| 101 | 6 |
| 110 | 7 |
| 111 | 8 |

**Table 3.2: Encoding/Decoding of the 1 binary decimal weight.**

| Genotype | Phenotype |
|----------|-----------|
| 000 | 2 |
| 001 | 1.5 |
| 010 | 1 |
| 011 | 0.5 |
| 100 | 0 |
| 101 | -0.5 |
| 110 | -1 |
| 111 | -1.5 |

### 3.3.2. The 3bit integer encoding/decoding scheme

The encoding/decoding of the delay times of the synapses can be seen in Table 3.3, which is the same as in the previous coding scheme. Finally, the coding of the weights can be observed in Table 3.4.

Table 3.3: Encoding/Decoding of the synapse delay times in ms.

| Genotype | Phenotype |
|---|---|
| 000 | 1 |
| 001 | 2 |
| 010 | 3 |
| 011 | 4 |
| 100 | 5 |
| 101 | 6 |
| 110 | 7 |
| 111 | 8 |

Table 3.4: Encoding/Decoding of the integer weight.

| Genotype | Phenotype |
|---|---|
| 000 | 4 |
| 001 | 3 |
| 010 | 2 |
| 011 | 1 |
| 100 | 0 |
| 101 | -1 |
| 110 | -2 |
| 111 | -3 |

### 3.3.3. The structure of a chromosome

As can be seen in Figure 3.4, each synapse is described by 6 bits in total. The first 3 bits represent the delay of the synapse and the latter 3 bits are the value of the synaptic weight. Placing all the encoded synapses next to each other, layer by layer, chromosome is formed. As an example, the total size of the chromosome in Figure 3.4 is 18 bits.

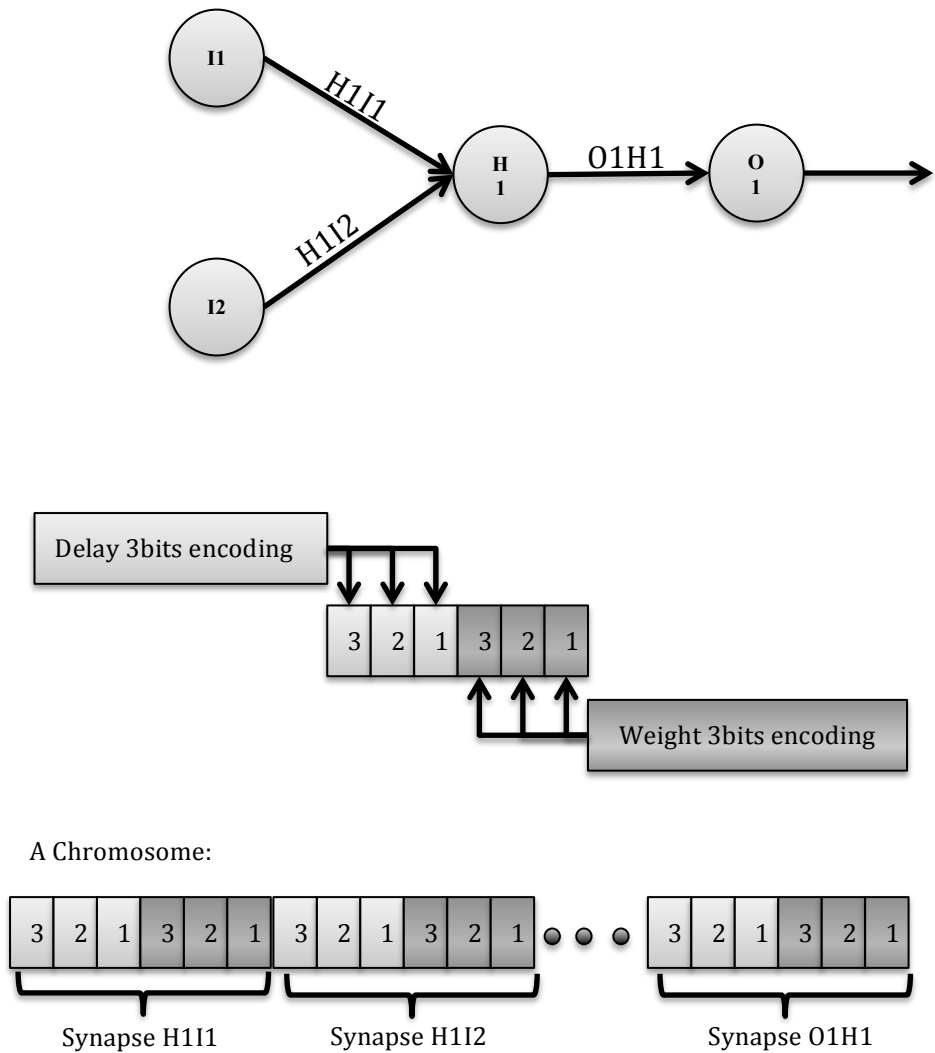

**Figure 3.4: The formation of a chromosome.**

# 4. The proposed algorithm for the XOR classification problem

## 4.1. Introduction

The XOR classification problem is often used as a first benchmark for a spiking neural network supervised learning algorithm due to the small size of its dataset. The truth table of the XOR gate can be seen in Table 4.1.

Table 4.1: The truth table of the XOR gate.

| A | B | OUT |
|---|---|-----|
| 0 | 0 | **0** |
| 0 | 1 | **1** |
| 1 | 0 | **1** |
| 1 | 1 | **0** |

Since information in spiking neural networks is coded as spikes, the XOR dataset has to be encoded as spike-times prior to any training.

The algorithm was tested on two different limited precision schemes, for three different network architectures and for two different simulation time steps. The same initial population was used for comparison reasons.

## 4.2. Encoding of the XOR problem into spike-times

Bohte et al. [14] proposed a very simple way to encode the XOR dataset into spike-times. For the inputs A and B, an input spike at 0ms would represent logic 0 while a spike at 6ms would represent logic 1 and for the output, a spike at 16ms would represent logic 0 while a spike at 10ms would represent logic 1. These values were chosen based on the trial-and-error method.

For this thesis the same method was used, with a small modification since in the spiking neural network program described in chapter 2, a spike 0 means no spike at all. The encoding of the XOR problem can be seen in Table 4.2.

Table 4.2: The XOR problem encoded into spike-times.

| Input Neuron 1 (ms) Bias neuron | Input Neuron 2 (ms) | Input Neuron 3 (ms) | Output Neuron (ms) |
|---|---|---|---|
| 1 | 1 | 1 | 17 |
| 1 | 1 | 7 | 10 |
| 1 | 7 | 1 | 10 |
| 1 | 7 | 7 | 17 |

An additional neuron, the bias neuron, is needed because without it the input spike patterns {1ms 1ms} and {7ms 7ms} would be treated as being the same.

## 4.3. Solving the XOR problem using the proposed algorithm with the one binary decimal place weight precision scheme

### 4.3.1. Network architecture 3 5 1

In Figure 4.1 the proposed spiking neural network architecture from Gosh et al.[43] can be seen. They used the same architecture to compare three supervised training algorithms: Spikeprop, QuickProp and RProp.

Similar architecture was used here, however, the biggest difference was that between every two neurons of different layers, only one synapse exists. In contrast, the aforementioned supervised training algorithms use a number of sub-synapses, which increases the computations since each neuron has to calculate all the incoming postsynaptic potentials.

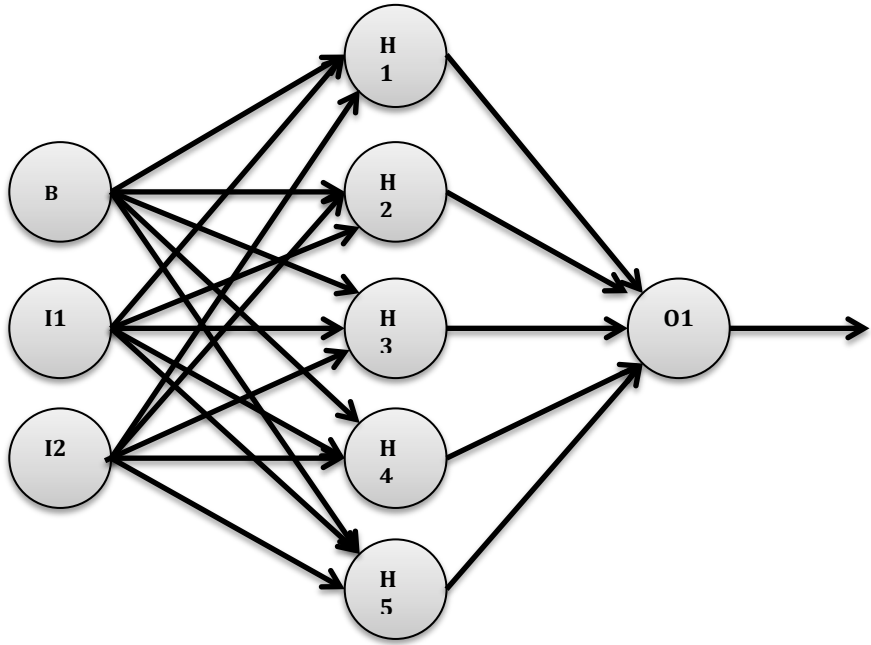

Figure 4.1: A spiking neural network architecture proposed by Gosh et al[43].

This architecture has 3 input neurons, one of which is the bias, 1 hidden layer with 5 neurons and 1 output layer with 1 neuron. So the topology is expressed as Topology=[3 5 1].

The settings of the spiking neural network and genetic algorithm can be seen in Tables 4.3 & 4.4. These genetic algorithm parameters were chosen because they produced the best results after a series of tests with the same initial population. The same settings were applied to all the trainings of this chapter.

The training algorithm was tested for different simulation time steps. The results can be seen in the following sections.

Table 4.3: Spiking neural network settings

| Topology | [3 5 1] |
|---|---|
| Simulation time | 50ms |
| Tau | 3 |
| tauR | 20 |
| Threshold | 1.5 |
| Maximum spikes | 10 |

Table 4.4: Genetic algorithm settings

| Bits size of the weights | 3 |
|---|---|
| Bits size of the delay times | 3 |
| Crossover rate | 0.6 |
| Mutation rate | 0.01 |
| Selective pressure | 1.5 |
| Elitism operator | 8 |
| Population size | 200 |

### *4.3.1.1. Simulation time step 0.01*

A time step of 0.01 was chosen for this training. The algorithm converged after 40 generations to a mean-squared error (MSE) of 0.09505. The time needed for a generation was 7.4 minutes.

The trained synaptic weights and delays can be seen in Tables 4.5 and 4.6, while in Figure 4.2 is their histogram. In Figure 4.3 the membrane potential of the neurons can be observed for the input spike patterns of the XOR problem. Finally, the plot of the generations versus best and average mean squared error can be seen in Appendix E.1, Figure E. 1.

**Table 4.5: Trained weights and delays of the synapses between the hidden and input layer. The following format is used (weight, delay).**

|    | B      | I1    | I2     |
|----|--------|-------|--------|
| H1 | 0, 2   | -1, 5 | 2, 6   |
| H2 | 2, 8   | 0, 5  | -0.5, 4|
| H3 | 1.5, 8 | 2, 2  | -1.5, 1|
| H4 | -1.5, 7| 1.5, 8| -1.5, 2|
| H5 | 2, 4   | -0.5, 3| 1, 1  |

**Table 4.6: Trained weights and delays of the synapses between the output and hidden layer. The following format is used (weight, delay).**

|    | H1     | H2   | H3     | H4      | H5     |
|----|--------|------|--------|---------|--------|
| O1 | 1.5, 1 | 1, 7 | 1.5, 5 | -1.5, 5 | 0.5, 1 |

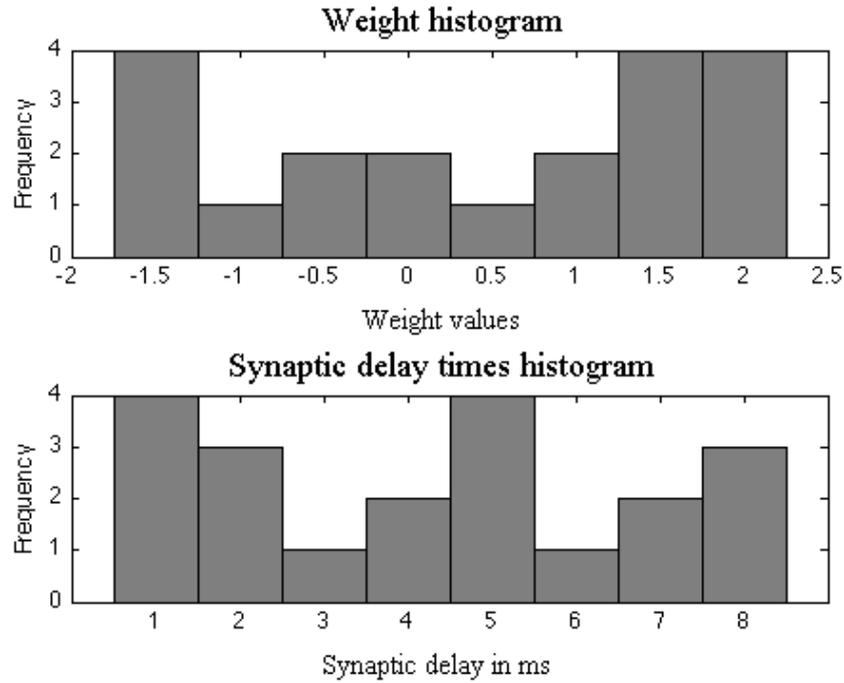

**Figure 4.2: The histogram of the trained synapses.**

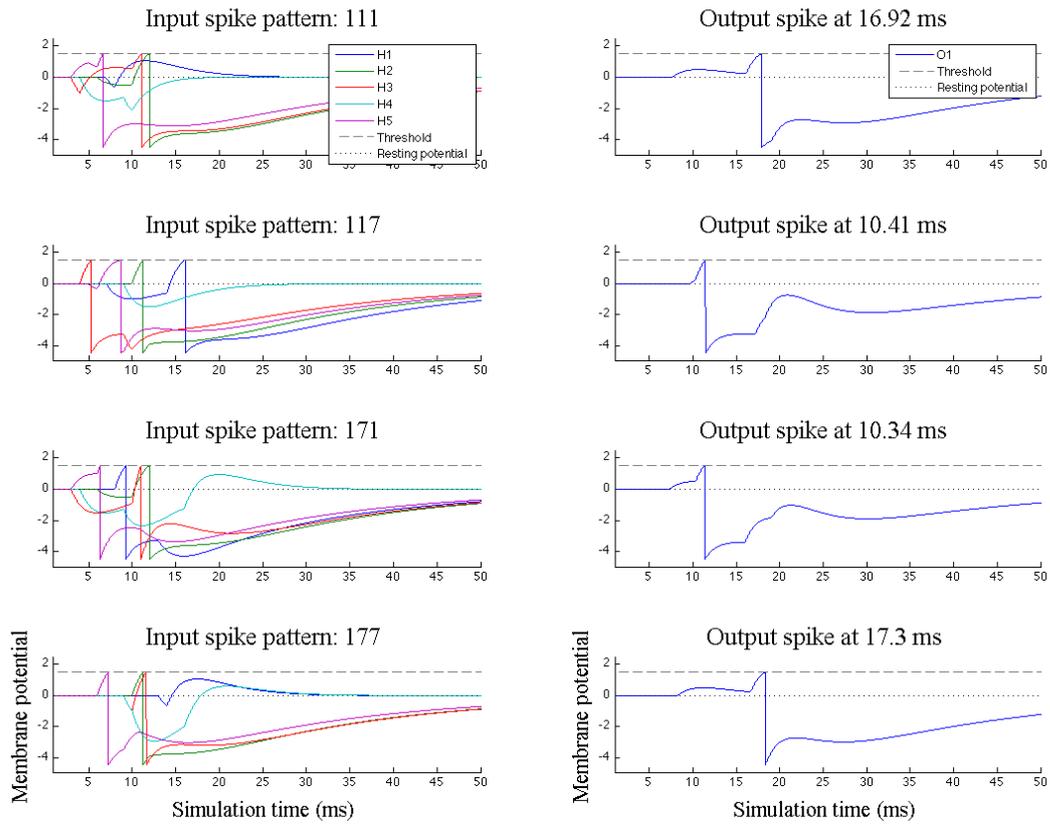

**Figure 4.3: The membrane potential of the neurons in the hidden layer (plots on the left hand site) and in the output layer (plots on the right hand site)**

### *4.3.1.2. Simulation time step 1*

For this training the simulation step of the spiking neural network was set to 1. The algorithm converged after 87 generations to a mean squared error (MSE) of 0. The time needed for a generation was 0.08 minutes.

The trained synaptic weights and delays can be seen in Tables 4.7 and 4.8, while in Figure 4.4 is their histogram. Finally, in Figure 4.5 the membrane potential of the neurons can be observed for the input spike patterns of the XOR problem. The plot of the generations versus best and average mean squared error can be seen in Appendix E.1, Figure E. 2.

**Table 4.7: Trained weights and delays of the synapses between the hidden and input layer. The following format is used (weight, delay).**

|     | B    | I1    | I2     |
|-----|------|-------|--------|
| H1  | 0, 6 | -1, 6 | 2, 5   |
| H2  | -1, 8| -1, 4 | 1.5, 2 |
| H3  | 2, 8 | 1.5, 1| -1, 1  |
| H4  | 1, 3 | 1.5, 4| -1.5, 3|
| H5  | 2, 4 | 0, 8  | -1, 7  |

**Table 4.8: Trained weights and delays of the synapses between the output and hidden layer. The following format is used (weight, delay).**

|     | H1   | H2   | H3   | H4     | H5     |
|-----|------|------|------|--------|--------|
| O1  | 2, 1 | 1, 1 | 2, 6 | 1.5, 2 | 0.5, 1 |

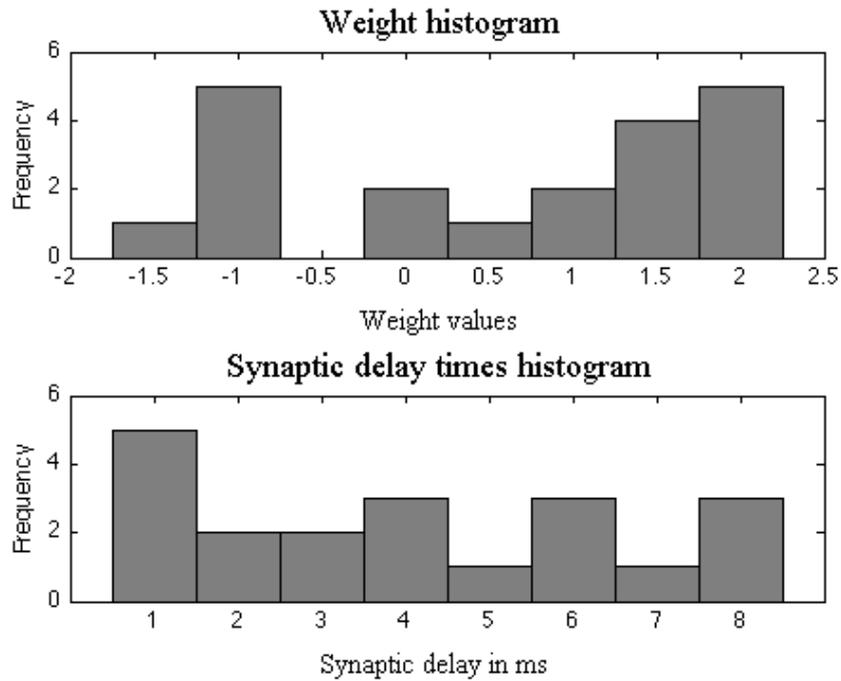

**Figure 4.4: The histogram of the trained synapses.**

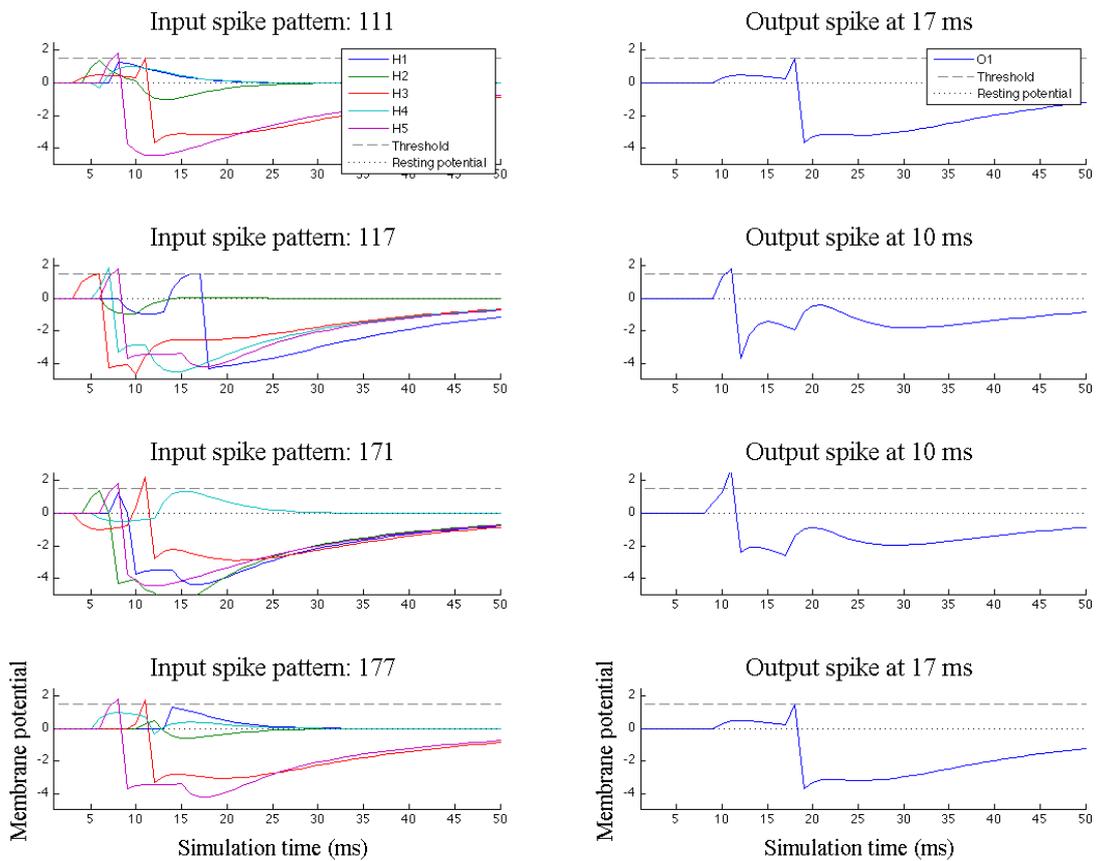

**Figure 4.5: The membrane potential of the neurons in the hidden layer (plots on the left hand site) and in the output layer (plots on the right hand site).**

### 4.3.2. Network architecture 3 2 1

This time the training algorithm was tested on smaller network architecture than the one that was proposed in [14,20,43]. The new architecture can be seen in Figure 4.6.

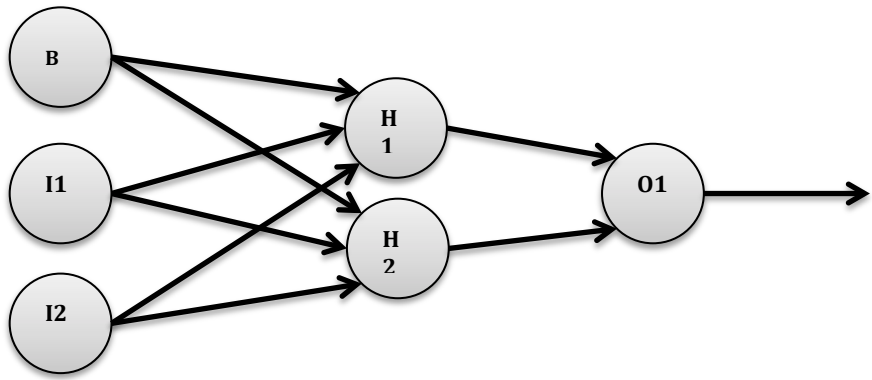

**Figure 4.6: A Spiking neural network with a smaller architecture than the proposed one.**

The Topology variable becomes Topology=[3 2 1] and the new spiking neural network settings can be seen in Table 4.9. The same genetic algorithm settings were used as shown in Table 4.4. The results can be seen in the following sections.

**Table 4.9: Spiking neural network settings**

| Topology | [3 2 1] |
|---|---|
| Simulation time | 50ms |
| Tau | 3 |
| tauR | 20 |
| Threshold | 1.5 |
| Maximum spikes | 10 |

#### 4.3.2.1. Simulation time step 0.01

The simulation time step was set to 0.01 and the training algorithm converged after 413 generations to a mean square error (MSE) of 0.2501. The time needed for a generation was 3.4 minutes.

The trained synaptic weights and delays can be seen in Tables 4.10 and 4.11, while in Figure 4.7 is their histogram. Finally, in Figure 4.8 the membrane potential of the neurons can be observed for the input spike patterns of the XOR problem. The plot of the generations versus best and average mean squared error can be seen in Appendix E.1, Figure E. 3.

Table 4.10: Trained weights and delays of the synapses between the hidden and input layer. The following format is used (weight, delay).

|    | B      | I1     | I2      |
|----|--------|--------|---------|
| H1 | 0.5, 2 | -1, 1  | 1, 2    |
| H2 | 1.5, 7 | 1.5, 2 | -1.5, 2 |

Table 4.11: Trained weights and delays of the synapses between the output and hidden layer. The following format is used (weight, delay).

|    | H1     | H2     |
|----|--------|--------|
| O1 | 1.5, 1 | 1.5, 2 |

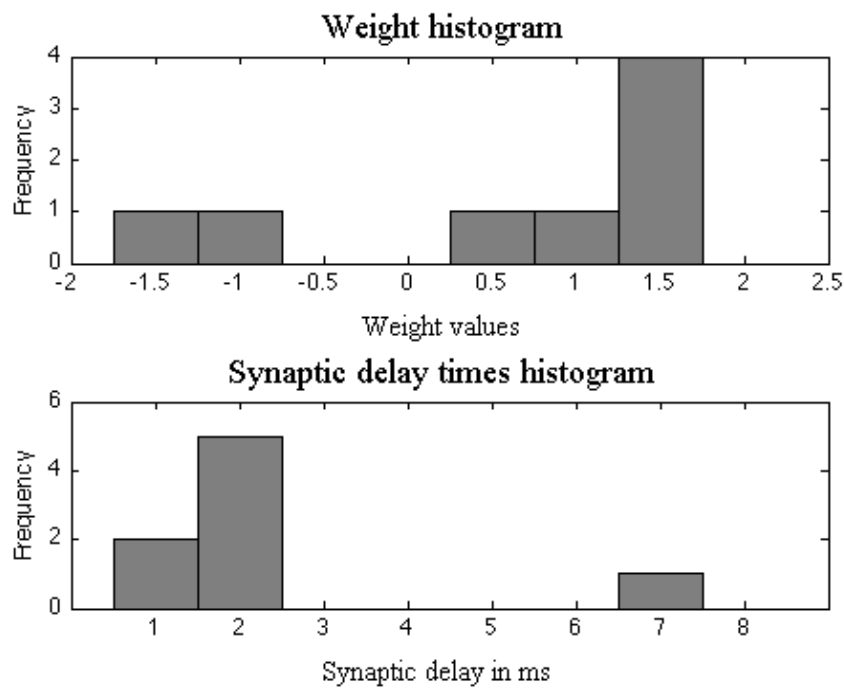

Figure 4.7: The histogram of the trained synapses.

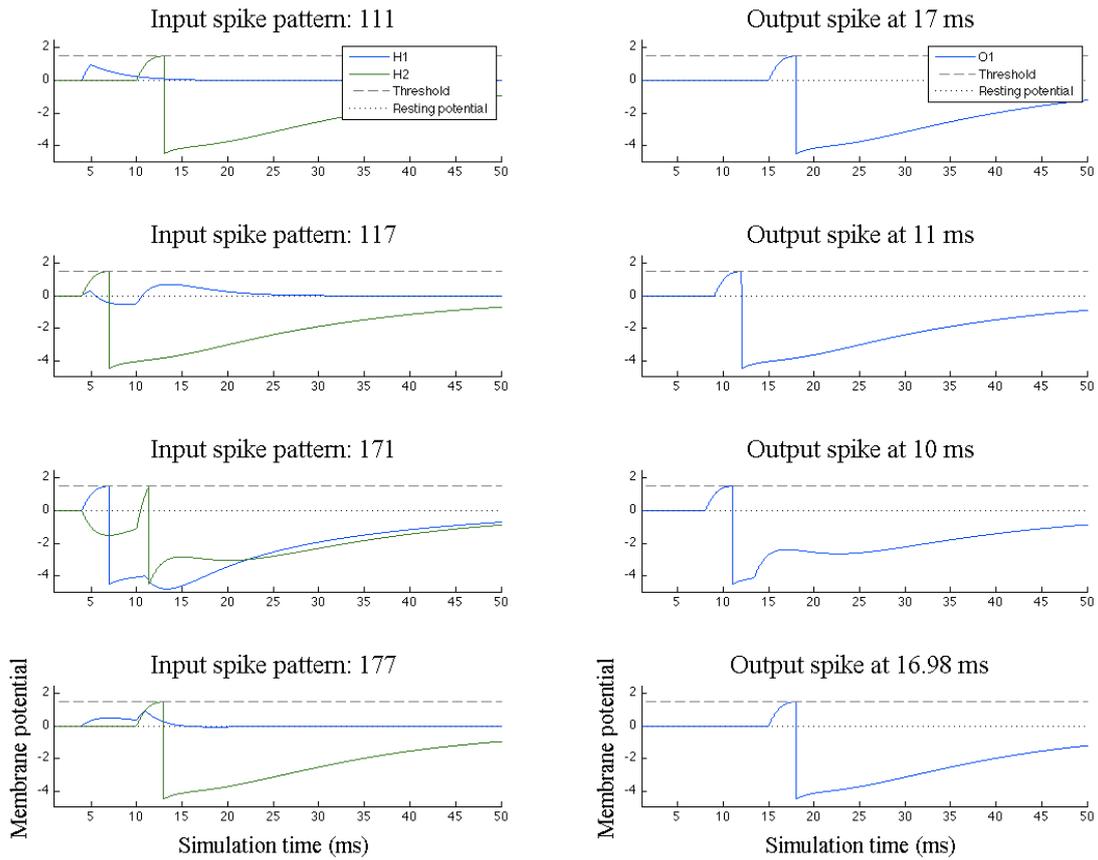

**Figure 4.8: The membrane potential of the neurons in the hidden layer (plots on the left hand site) and in the output layer (plots on the right hand site)**

### *4.3.2.2. Simulation time step 1*

This time the simulation time step was set to 1 and the training algorithm converged after 64 generations to a mean squared error (MSE) of 0.25. The time needed for a generation was 0.04 minutes.

The trained synaptic weights and delays can be seen in Tables 4.12 and 4.13, while in Figure 4.9 is their histogram. Finally, in Figure 4.10 the membrane potential of the neurons can be observed for the input spike patterns of the XOR problem. The plot of the generations versus best and average mean squared error can be seen in Appendix E.1, Figure E. 4.

Table 4.12: Trained weights and delays of the synapses between the hidden and input layer. The following format is used (weight, delay).

|    | B    | I1   | I2     |
|----|------|------|--------|
| H1 | 0, 2 | -1, 1| 2, 4   |
| H2 | 1.5, 8 | 2, 1 | -1.5, 1 |

Table 4.13: Trained weights and delays of the synapses between the output and hidden layer. The following format is used (weight, delay).

|    | H1   | H2   |
|----|------|------|
| O1 | 2, 1 | 2, 4 |

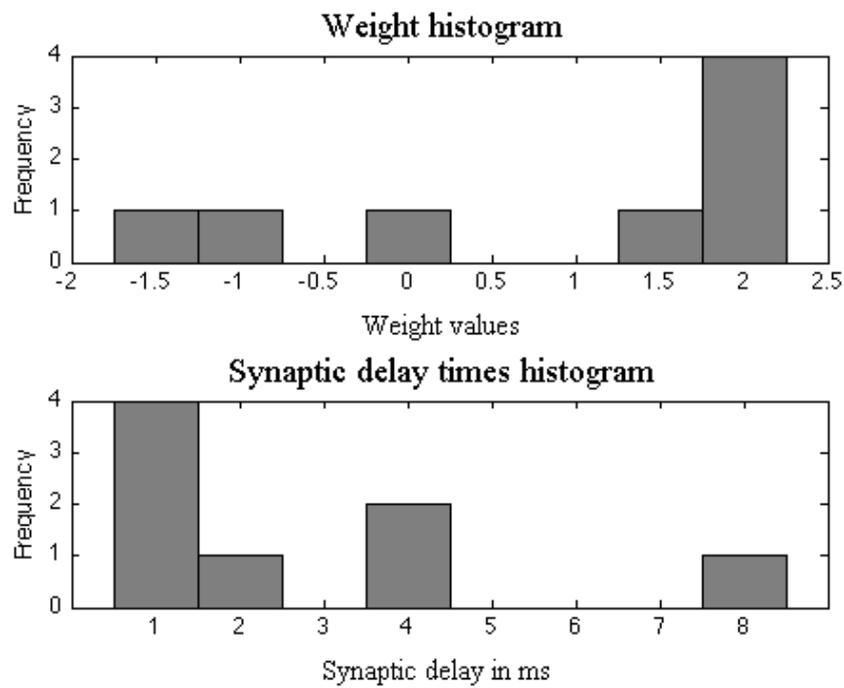

Figure 4.9: The histogram of the trained synapses.

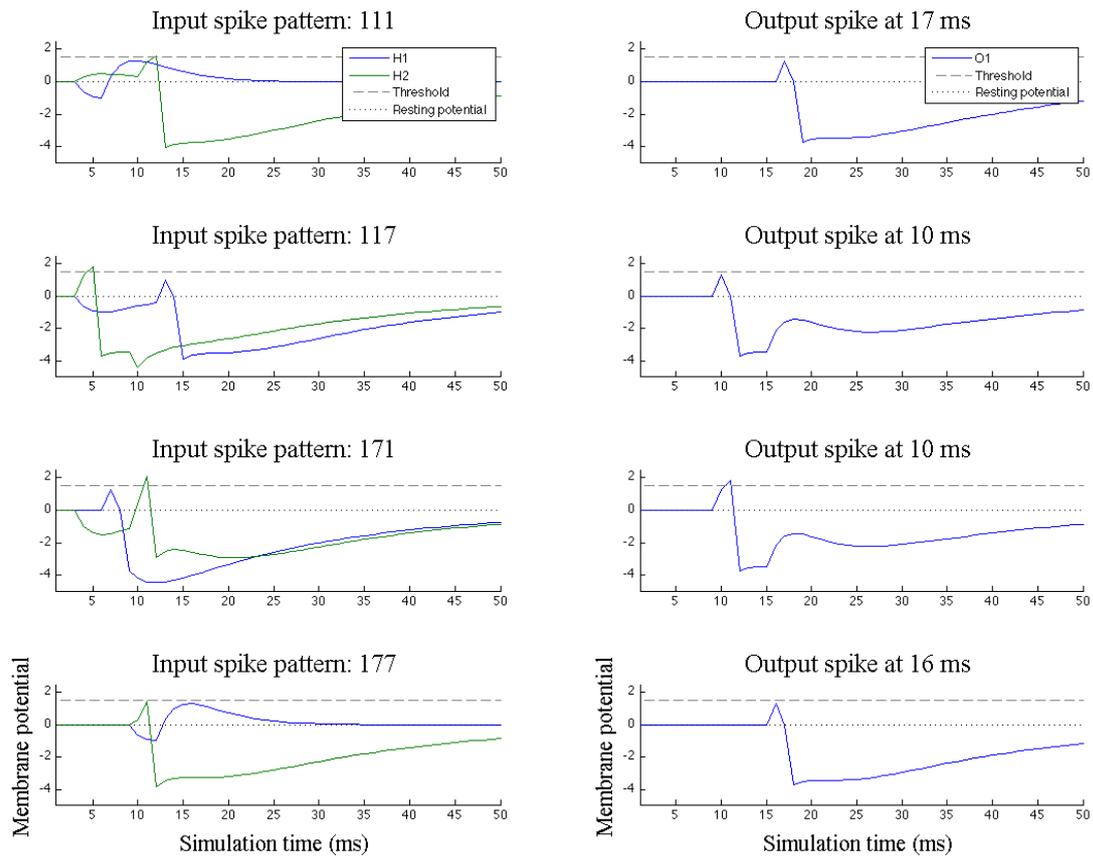

Figure 4.10: The membrane potential of the neurons in the hidden layer (plots on the left hand site) and in the output layer (plots on the right hand site)

## 4.4. Solving the XOR problem using the proposed algorithm with the integer weight precision scheme

In this the section the same trainings were done but this time the integer-coding scheme, as described in chapter 3.3.2, was used.

### 4.4.1. Network architecture 3 5 1

The spiking neural network architecture can be observed in Figure 4.1. Furthermore, the same simulation and genetic algorithm settings were applied and can be seen in Tables 4.3 & 4.4.

#### 4.4.1.1. Simulation time step 0.01

The simulation time step was set to 0.01 and the algorithm was able to converge after 21 generations to a mean squared error (MSE) of 0.07135. The time needed for a generation was 7.3 minutes.

The trained synaptic weights and delays can be seen in Tables 4.14 and 4.15, while in Figure 4.11 is their histogram. Finally, in Figure 4.12 the membrane potential of the neurons can be observed for the input spike patterns of the XOR problem. The plot of the generations versus best and average mean squared error can be seen in Appendix E.1, Figure E. 5.

Table 4.14: Trained weights and delays of the synapses between the hidden and input layer. The following format is used (weight, delay).

|    | B    | I1   | I2   |
|----|------|------|------|
| H1 | 0, 6 | -2, 5 | 4, 6 |
| H2 | 4, 8 | 0, 6 | 0, 2 |
| H3 | 3, 8 | 4, 1 | -3, 1 |
| H4 | -3, 8 | 3, 8 | -3, 1 |
| H5 | 4, 2 | 0, 3 | -2, 6 |

Table 4.15: Trained weights and delays of the synapses between the output and hidden layer. The following format is used (weight, delay).

|    | H1   | H2   | H3   | H4   | H5   |
|----|------|------|------|------|------|
| O1 | 3, 1 | 4, 7 | 4, 6 | -3, 1 | 1, 3 |

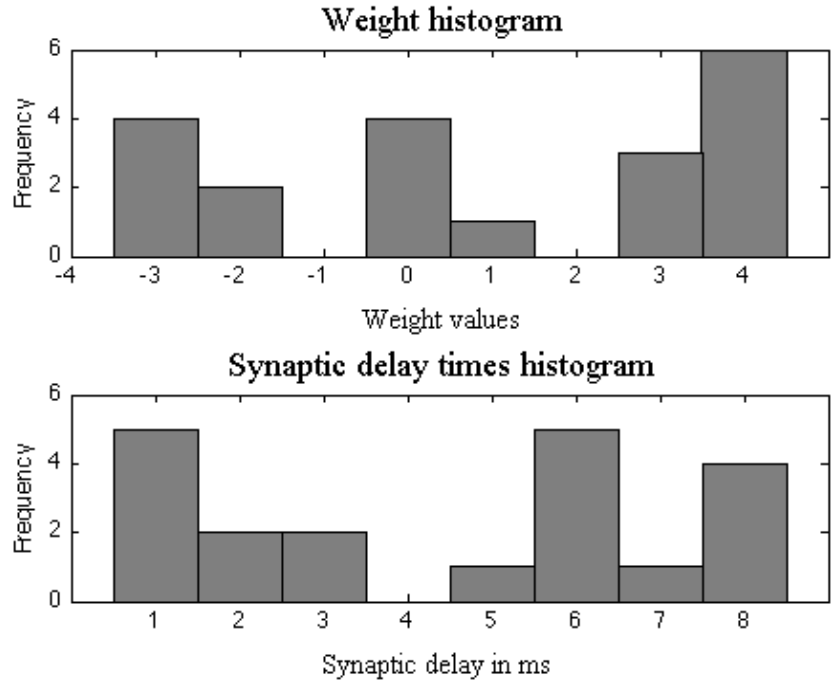

**Figure 4.11: The histogram of the trained synapses.**

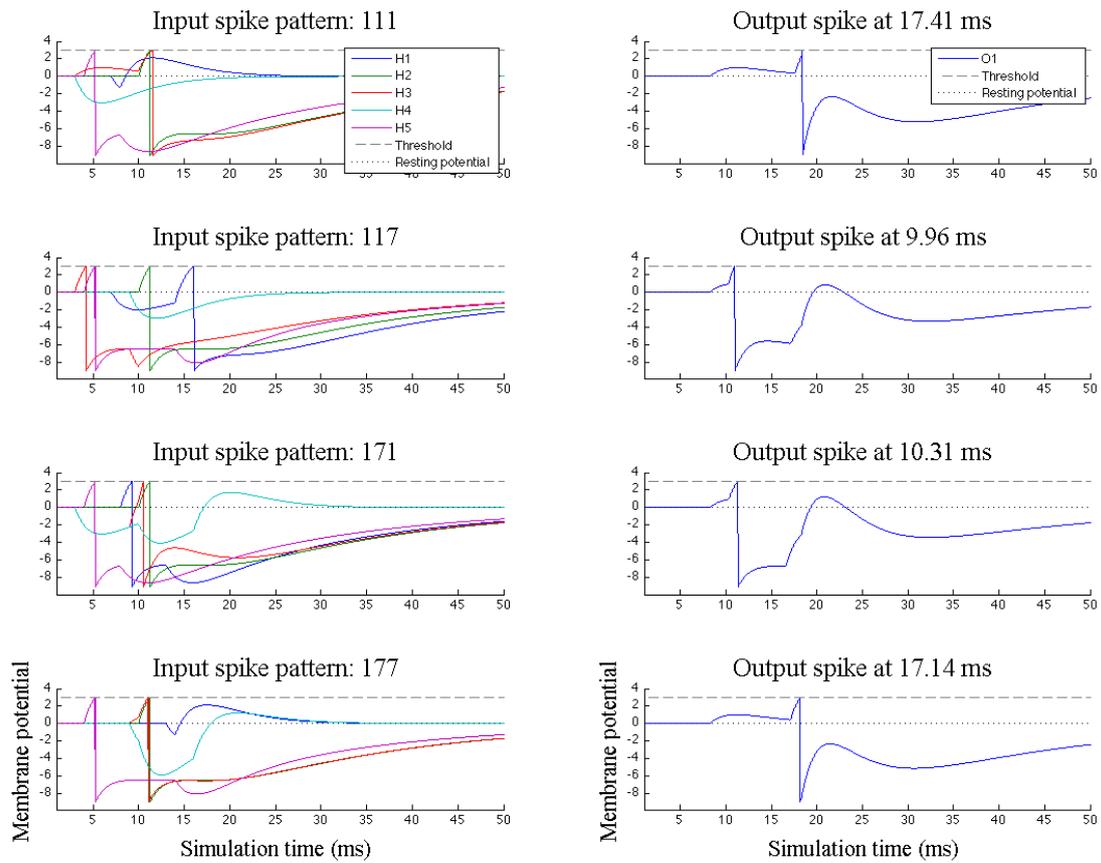

**Figure 4.12: The membrane potential of the neurons in the hidden layer (plots on the left hand site) and in the output layer (plots on the right hand site)**

*4.4.1.2. Simulation time step 1*

For this training the simulation time step was set to 1 and the training algorithm converged after 143 generations to a mean squared error (MSE) of 0. The time needed for a generation was 0.08 minutes.

The trained synaptic weights and delays can be seen in Tables 4.16 and 4.17, while in Figure 4.13 is their histogram. Finally, in Figure 4.14 the membrane potential of the neurons can be observed for the input spike patterns of the XOR problem. The plot of the generations versus best and average mean squared error can be seen in Appendix E.1, Figure E. 6.

**Table 4.16: Trained weights and delays of the synapses between the hidden and input layer. The following format is used (weight, delay).**

|    | B    | I1    | I2    |
|----|------|-------|-------|
| H1 | 0, 1 | -2, 6 | 4, 5  |
| H2 | 3, 7 | -1, 8 | 0, 8  |
| H3 | 3, 8 | 4, 1  | -3, 1 |
| H4 | 2, 8 | 2, 1  | -3, 1 |
| H5 | 4, 4 | 2, 1  | -2, 7 |

**Table 4.17: Trained weights and delays of the synapses between the output and hidden layer. The following format is used (weight, delay).**

|    | H1   | H2    | H3   | H4   | H5   |
|----|------|-------|------|------|------|
| O1 | 3, 1 | -2, 2 | 4, 5 | 4, 4 | 2, 2 |

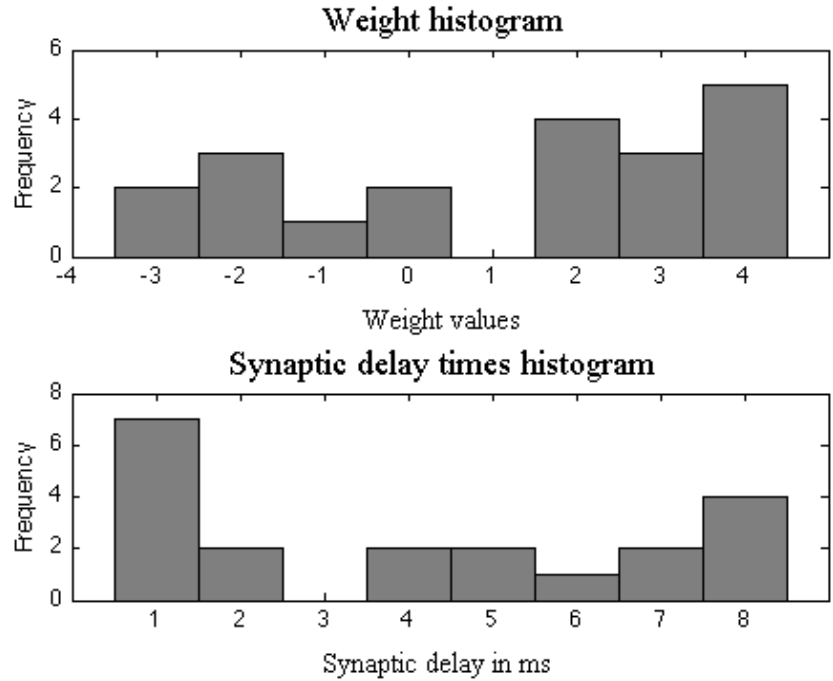

**Figure 4.13: The histogram of the trained synapses.**

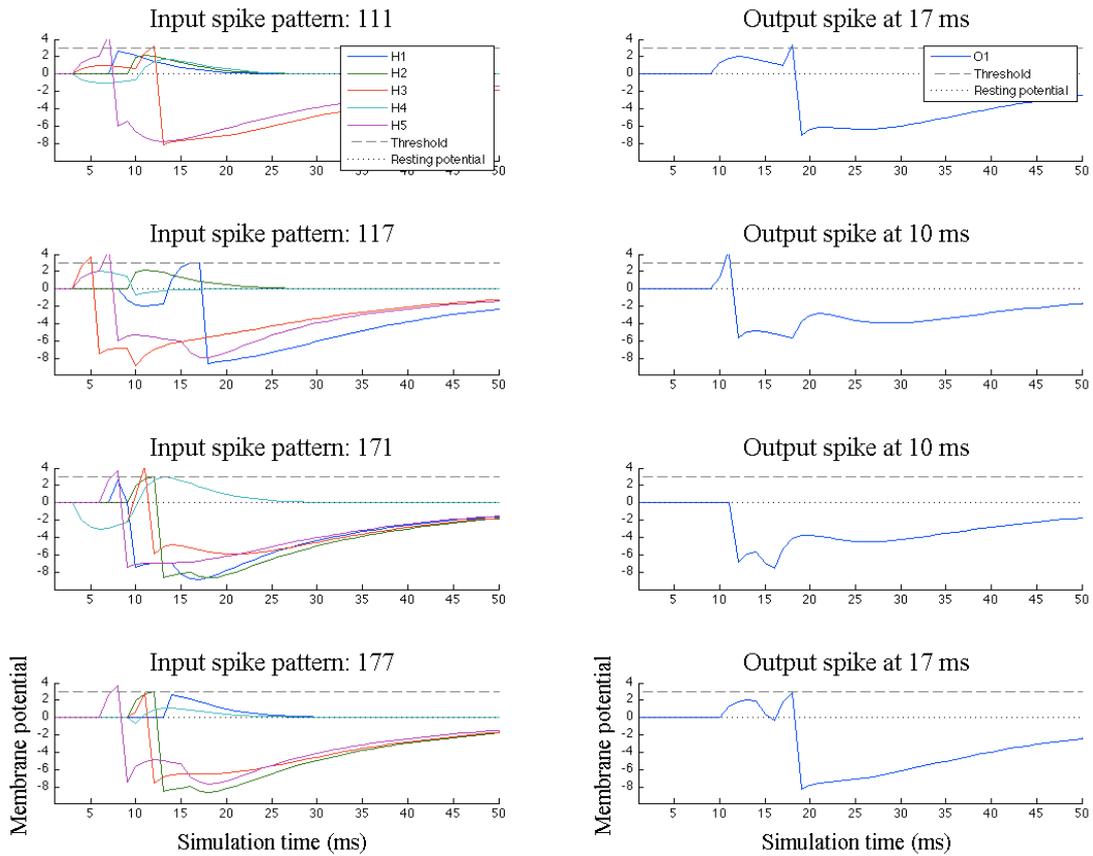

**Figure 4.14: The membrane potential of the neurons in the hidden layer (plots on the left hand site) and in the output layer (plots on the right hand site)**

### 4.4.2. Network architecture 3 2 1

The spiking neural network architecture that was used in this section is the same as in Figure 4.6. In addition, the simulation settings can be seen in Table 4.3, while the genetic algorithm settings can be observed in Table 4.4.

#### *4.4.2.1. Simulation time step 0.01*

The simulation time step was set to 0.01 and the training algorithm was able to converge after 157 generations to a mean squared error (MSE) of 0.112625. The time needed for a generation was 3.5 minutes.

The trained synaptic weights and delays can be seen in Tables 4.18 and 4.19, while in Figure 4.15 is their histogram. Finally, in Figure 4.16 the membrane potential of each neuron can be observed for the input spike patterns of the XOR problem. The plot of the generations versus best and average mean squared error can be seen in Appendix E.1, Figure E. 7.

**Table 4.18: Trained weights and delays of the synapses between the hidden and input layer. The following format is used (weight, delay).**

|    | B    | I1   | I2   |
|----|------|------|------|
| H1 | 3, 8 | -2, 1 | 4, 1 |
| H2 | 2, 7 | 4, 3 | -3, 3 |

**Table 4.19: Trained weights and delays of the synapses between the output and hidden layer. The following format is used (weight, delay).**

|    | H1   | H2   |
|----|------|------|
| O1 | 4, 6 | 3, 2 |

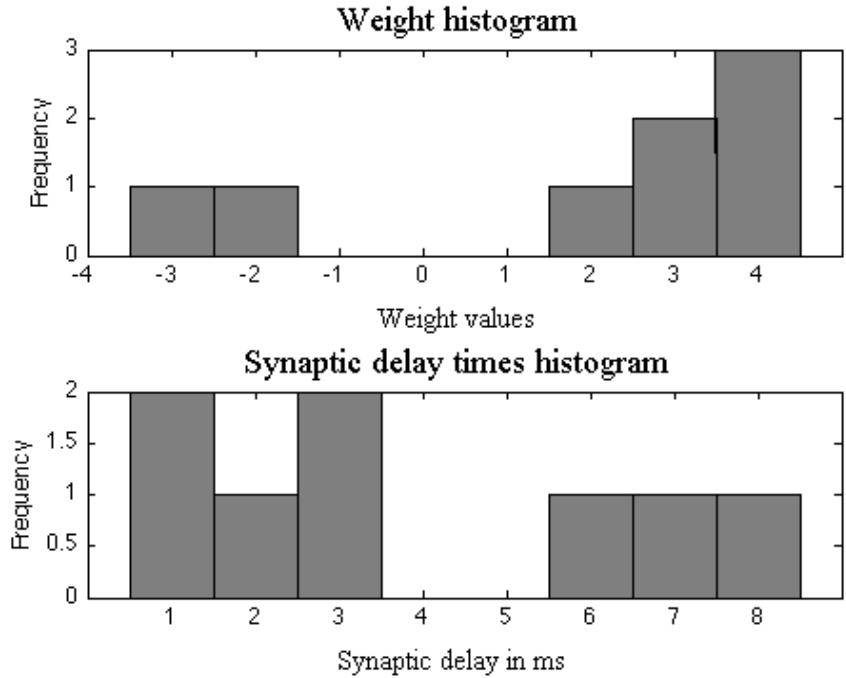

**Figure 4.15: The histogram of the trained synapses.**

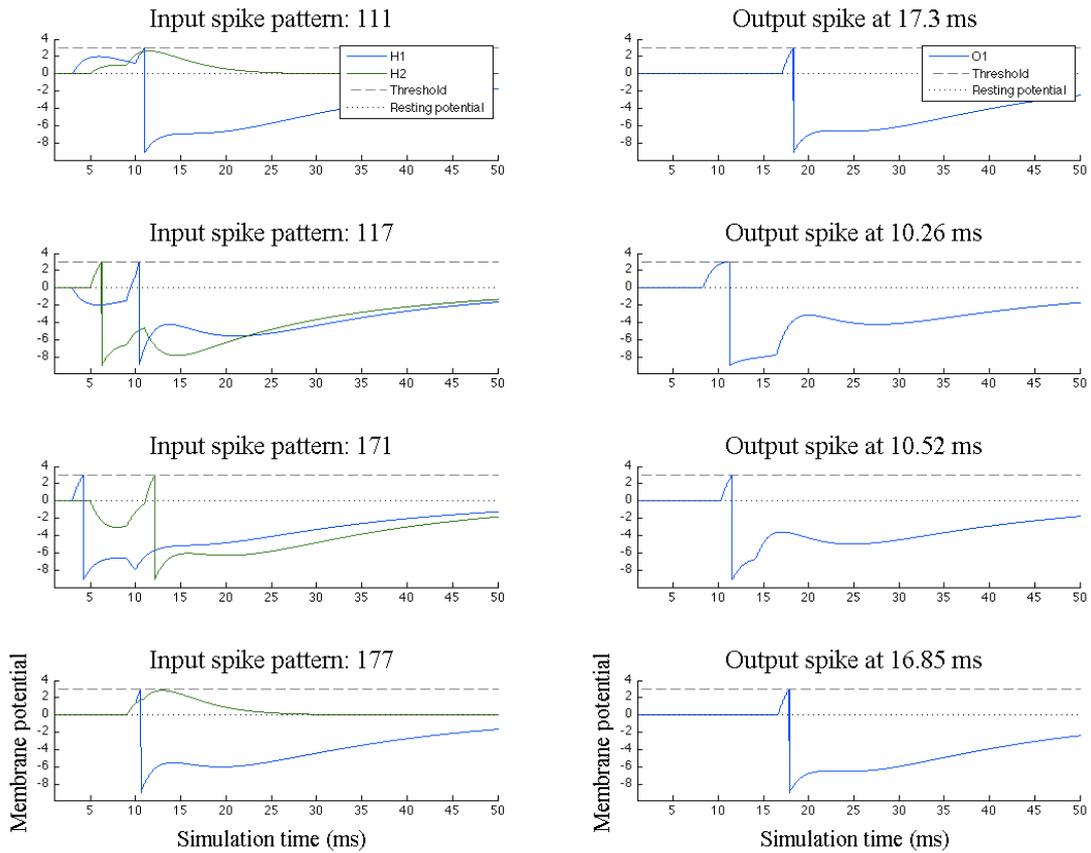

**Figure 4.16: The membrane potential of the neurons in the hidden layer (plots on the left hand site) and in the output layer (plots on the right hand site)**

*4.4.2.2. Simulation time step 1*

The simulation time step was set to 1 and the training algorithm converged after 285 generations to a mean squared error (MSE) of 0. The time needed for a generation was 0.04 minutes.

The trained synaptic weights and delays can be seen in Tables 4.20 and 4.21, while in Figure 4.17 is their histogram. Finally, in Figure 4.18 the membrane potential of each neuron can be observed for each input spike patterns of the XOR problem. The plot of the generations versus best and average mean squared error can be seen in Appendix E.1, Figure E. 8.

**Table 4.20: Trained weights and delays of the synapses between the hidden and input layer. The following format is used (weight, delay).**

|    | B    | I1   | I2   |
|----|------|------|------|
| H1 | 1, 1 | 4, 4 | -3, 1 |
| H2 | 3, 8 | -3, 1 | 4, 1 |

**Table 4.21: Trained weights and delays of the synapses between the output and hidden layer. The following format is used (weight, delay).**

|    | H1   | H2   |
|----|------|------|
| O1 | 3, 1 | 3, 3 |

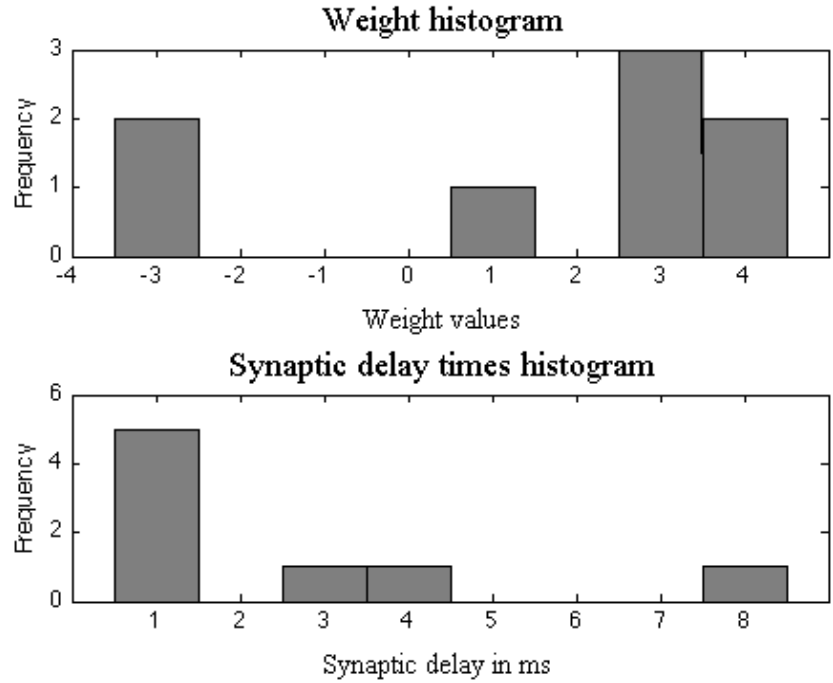

**Figure 4.17: The histogram of the trained synapses.**

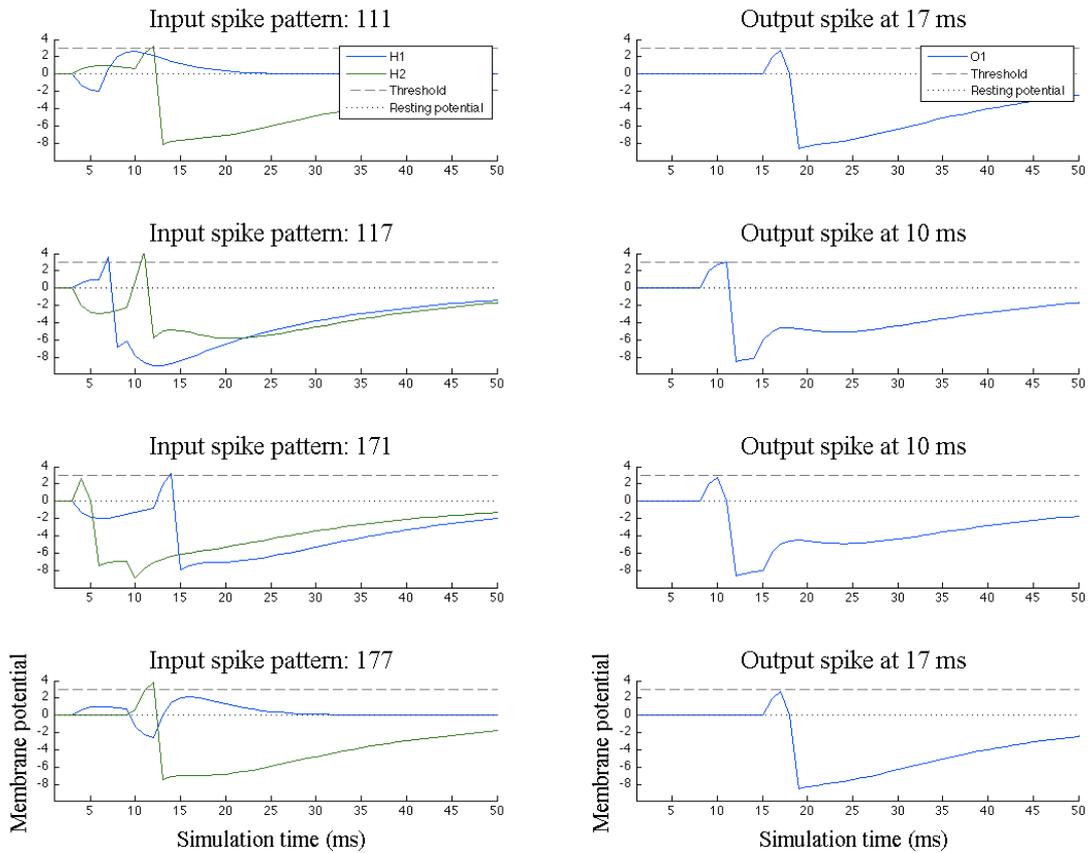

**Figure 4.18: The membrane potential of the neurons in the hidden layer (plots on the left hand site) and in the output layer (plots on the right hand site)**

## 4.5. Solving the XOR problem using one neuron

In this section the proposed training algorithm was tested on the XOR problem again but for an even smaller network. Only one neuron was used with the purpose of testing the ability of the algorithm to train both synaptic weights and delay times.

The XOR problem is a non-linear separable problem, Table 4.1, which means that a single perceptron cannot produce a solution. In order to solve the XOR problem using traditional artificial neural networks, a hidden layer is needed with 2 neurons [44], where each of these neurons produces a partial solution and then the output neuron combines those solutions and solves the XOR problem [38].

In this section, it will be shown that it is possible to solve this problem using only one spiking neuron. The proposed architecture can be seen in Figure 4.19.

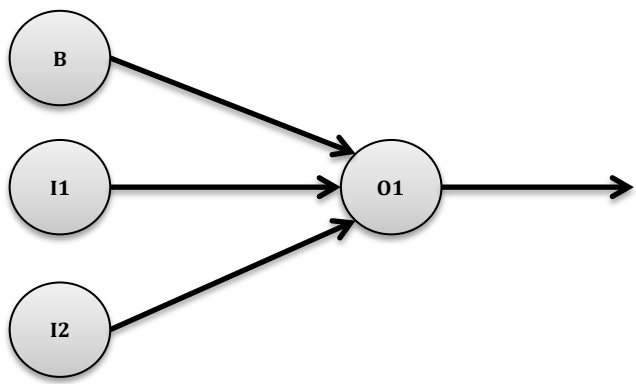

**Figure 4.19: The one neuron XOR. The input layer neurons do not count as they are non computing.**

However, a different input/output encoding was used this time. The encoding of the input is still spike-times but now a binary encoding is used for the output. The new encoding of the XOR problem can be seen in Table 4.22.

Table 4.22: The new XOR encoding.

| Input Neuron 1 (ms) B | Input Neuron 2 (ms) I1 | Input Neuron 3 (ms) I2 | Output Neuron (ms) O1 |
|---|---|---|---|
| 1 | 1 | 1 | No spike |
| 1 | 1 | 7 | 10 |
| 1 | 7 | 1 | 10 |
| 1 | 7 | 7 | No spike |

The spiking neural networks settings can be seen in Table 4.23 while the settings of the genetic algorithm can be observed in Table 4.4.

Table 4.23: The simulation settings for the one neuron XOR.

| Topology | [3 1] |
|---|---|
| Simulation time | 50ms |
| Simulation step | 1 |
| Tau | 3 |
| tauR | 20 |
| Threshold | 3 |
| Maximum spikes | 10 |

### 4.5.1. Using the one binary decimal place coding scheme

Using the one binary decimal place coding scheme, presented in section 3.3.1 and a simulation time step of 1 the training algorithm was able to converge, after 19 generations to a mean squared error (MSE) of 0.5. The time needed for a generation was 0.04 minutes.

The trained synaptic weights and delays can be seen in Table 4.24. Finally, in Figure 4.20 the membrane potential of the O1 neuron can be observed for the input spike patterns of the XOR problem. The plot of the generations versus best and average mean squared error can be seen in Appendix E.1, Figure E. 9.

**Table 4.24: Trained weights and delays of the synapses between the hidden and input layer. The following format is used (weight, delay).**

|     | B    | I1   | I2     |
|-----|------|------|--------|
| O1  | 2, 7 | 2, 2 | -1.5, 1 |

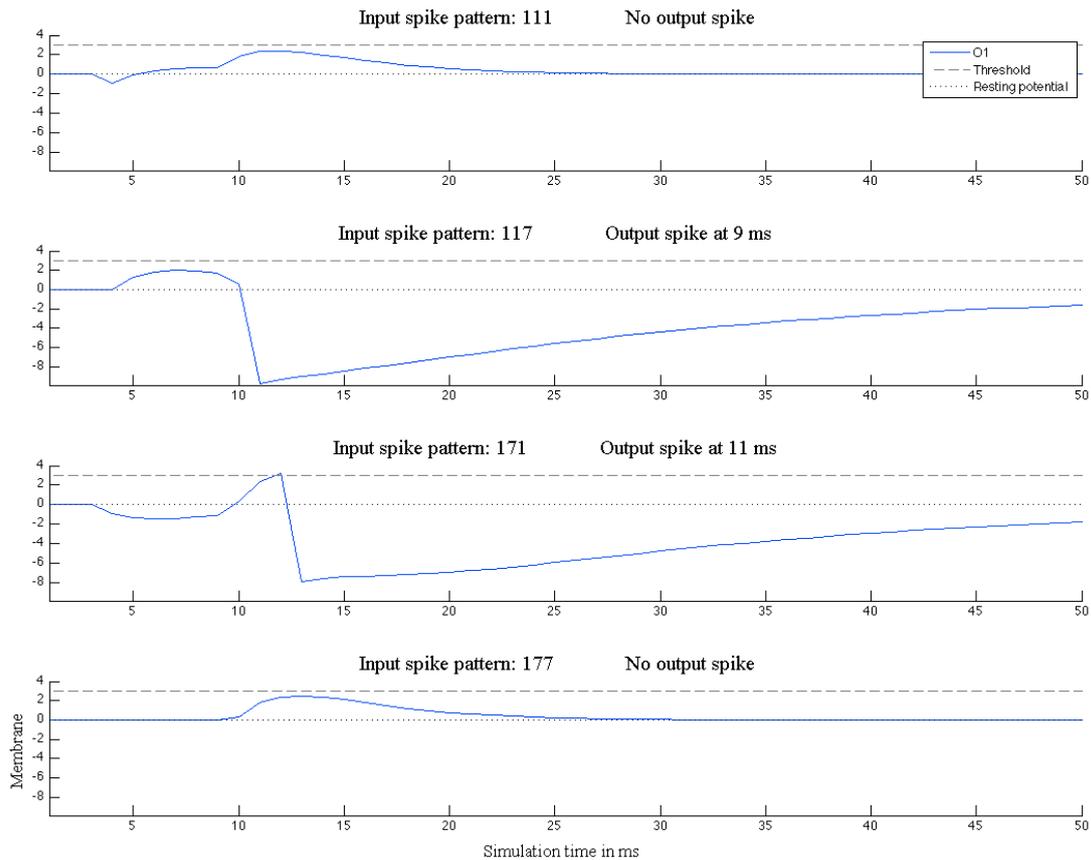

**Figure 4.20: The membrane potential of the neurons in the hidden layer (plots on the left hand site) and in the output layer (plots on the right hand site)**

### 4.5.2. Using the integer coding scheme

This time the same training was done using the integer coding scheme, as seen in section 3.3.2. The simulation time step was set to 1 and the training algorithm converged after 3 generations to a mean squared error (MSE) of 1. The time needed for a generation was 0.04 minutes.

The trained synaptic weights and delays can be seen in Table 4.25. Lastly, in Figure 4.21 the membrane potential of the O1 neuron can be observed for the input spike patterns of the XOR problem. The plot of the generations versus best and average mean squared error can be seen in Appendix E.1, Figure E. 10.

Table 4.25: Trained weights and delays of the synapses between the hidden and input layer. The following format is used (weight, delay).

|    | B   | I1   | I2  |
|----|-----|------|-----|
| O1 | 4, 8 | -3, 3 | 4, 3 |

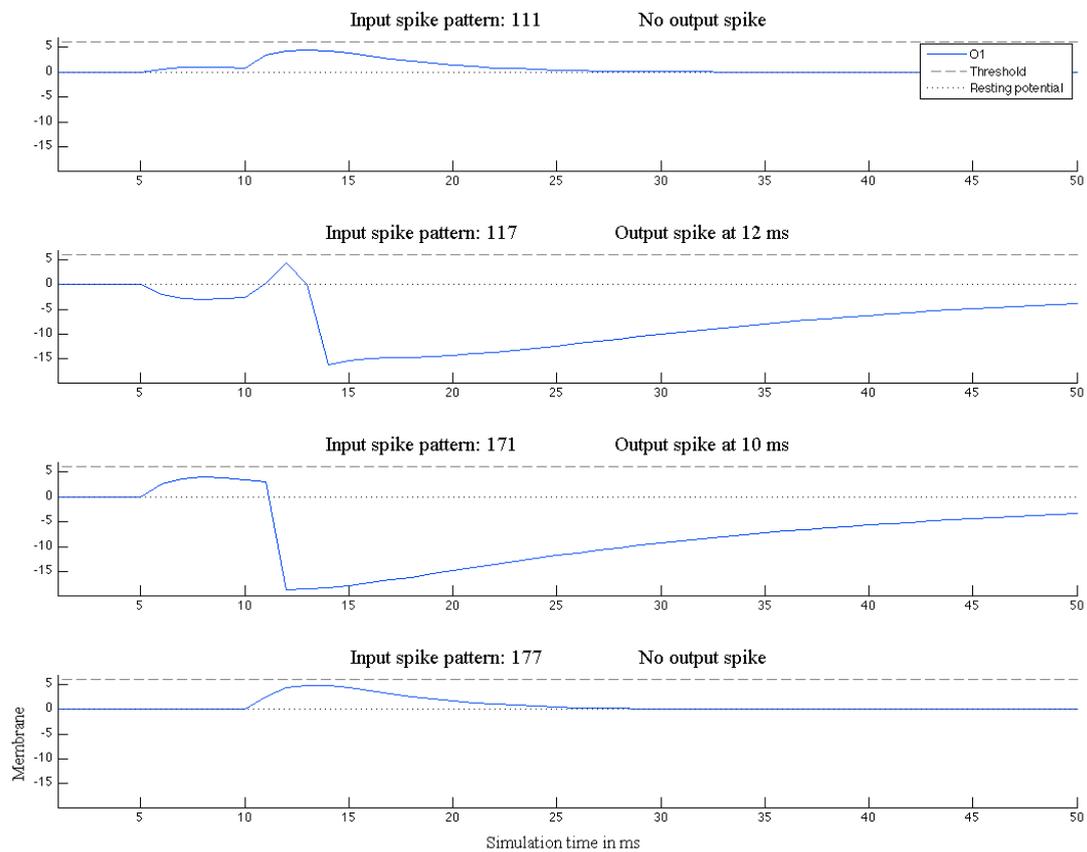

Figure 4.21: The membrane potential of the neurons in the hidden layer (plots on the left hand site) and in the output layer (plots on the right hand site).

### 4.6. Discussion on the training results

The training results of the XOR problem are summarized in Table 4.26. As it can be seen for the [3 5 1] network architecture, the proposed algorithm was able to converge to a much smaller mean squared error compared to SpikeProp, QuickProp, RProp [43]. Furthermore, the algorithm was able to train an even smaller network than the one that was used in [14,20,40,43].

In addition, the proposed algorithm was able to solve the XOR problem using only one neuron, something that is impossible for traditional artificial neural networks. This task was given as a test to prove the algorithm's power to train both synaptic weights and delays. However, as can be seen from Figure 4.21, in the cases of logic 0 the membrane potential of the O1 neuron reaches close to the threshold and if the neuron is implemented on analog hardware, then it might accidentally cross the threshold due to noise and emit a spike.

When comparing these results to the other supervised training algorithms, one should bare in mind that this algorithm is designed for single synapse per neuron whereas [14,20,28,43] use several sub-synapses per neuron, as seen in Figure 1.4. This means that the proposed algorithm needs fewer calculations for each neuron. Furthermore, this algorithm uses limited precision to train the synaptic weights and delays while the aforementioned ones use real values so, this makes it more suitable for hardware implementations.

Table 4.26: Summary of the XOR training results in terms of mean squared errors (MSE).

|                      | 1 binary decimal coding scheme | | Integer coding scheme | |
|----------------------|-----------------|-------------|-----------------|------------|
| **Network architecture** | *Time step 0.01* | *Time step 1* | *Time step 0.01* | *Time step1* |
| Topology = [3 5 1]   | 0.09505         | 0           | 0.07135         | 0          |
| Topology = [3 2 1]   | 0.2501          | 0.25        | 0.112625        | 0          |
| Topology = [3 1]     | -------------------- | 0.5    | -------------------- | 1     |

# 5. The proposed algorithm for the Fisher iris classification problem

## 5.1. Introduction

The Fisher iris dataset consists of three classes of iris flowers: Setosa, Versicolor, Virginica of 50 samples each and each of those classes has four attributes: Sepal Length, Sepal Width, Petal Length, Petal Width. Sir Ronald Aylmer Fisher created this data set in 1936 and it is one of the most widely used classification benchmarks for spiking neural networks.

The dataset was downloaded from the University of California, Irvine (UCI) [45] repository and can be seen in Appendix D.

## 5.2. Processing and plotting the dataset

In the Appendix B.3.1 is the Matlab source code of the function that opens the iris dataset and returns a structure file. In addition, there is one more function in Appendix B.3.2 that plots the Fisher structure file, Figure 5.1. As can be seen, the Setosa class is linearly separable to the other two classes.

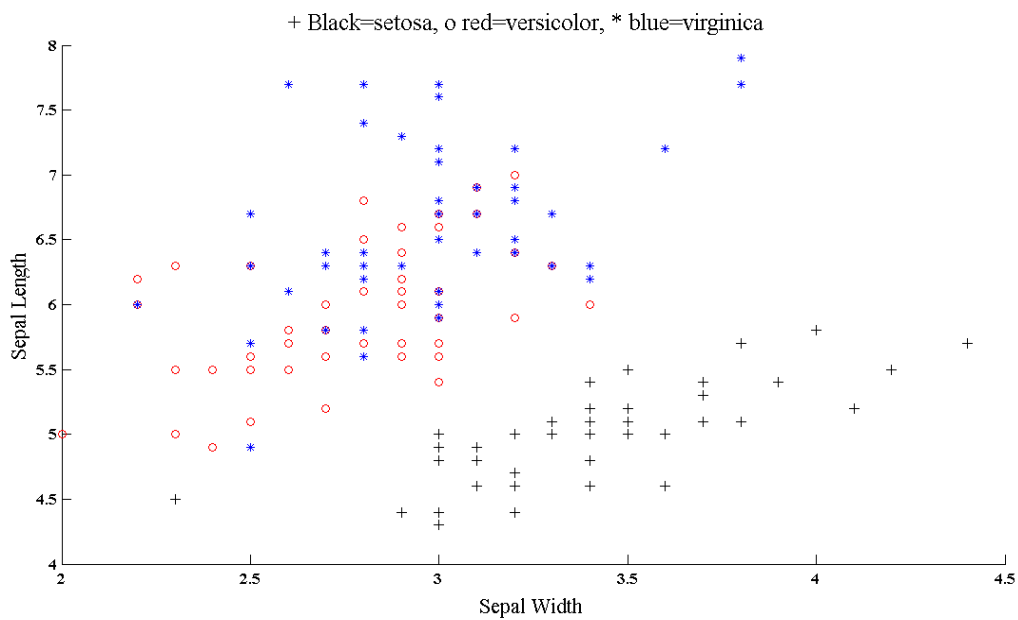

**Figure 5.1: The Fisher iris dataset.**

## 5.3. Converting the Fisher iris attributes into spike-times by using multiple Gaussian receptive fields

The real-valued features of the iris dataset have to be converted into spike-times. Since the data are not integer numbers, population coding has to be used as described in chapter 1.4.3. In this case 8 Gaussian Receptive Fields (GRF) for each input feature were used. At first, 4 GRF were tried out but proved not enough since the algorithm was not able to converge to a small mean squared error.

The centre and width of each Gaussian RF neuron was calculated using equations 1.19 and 1.20. The settings are summarized in Table 5.1. A value of γ=1.5 was suggested by references [11, 25, 43].

Table 5.1: Gaussian receptive field neuron settings

| m Gaussian RF neurons | 8 |
|---|---|
| Imin | 0 |
| Imax | 50 |
| γ | 1.5 |
| Threshold fire line | 0.1 |

In Figure 5.2, it can be seen how 8 Gaussian RF neurons, I1 to I8, convert a real number into spike-times. The x-axis represents the real numbers that are going to be expressed as spike-times. In this example, the number that is going to be converted into spike-time is the number 3.2. The red vertical line projects to the y-axis every time there is an intersection between the red line (real number) and a Gaussian RF neuron. Finally, the chosen values of the y-axis are multiplied by 10, then rounded to the time step and subtracted by 10 to generate a spike of that particular Gaussian RF neuron.

In the example of Figure 5.2 the I1 GRF neuron will emit a spike at 8ms, I2 GRF neuron at 1ms, I3 GRF at 5ms and finally I4 GRF neuron will not emit a spike because the point of intersection is under the firing threshold line.

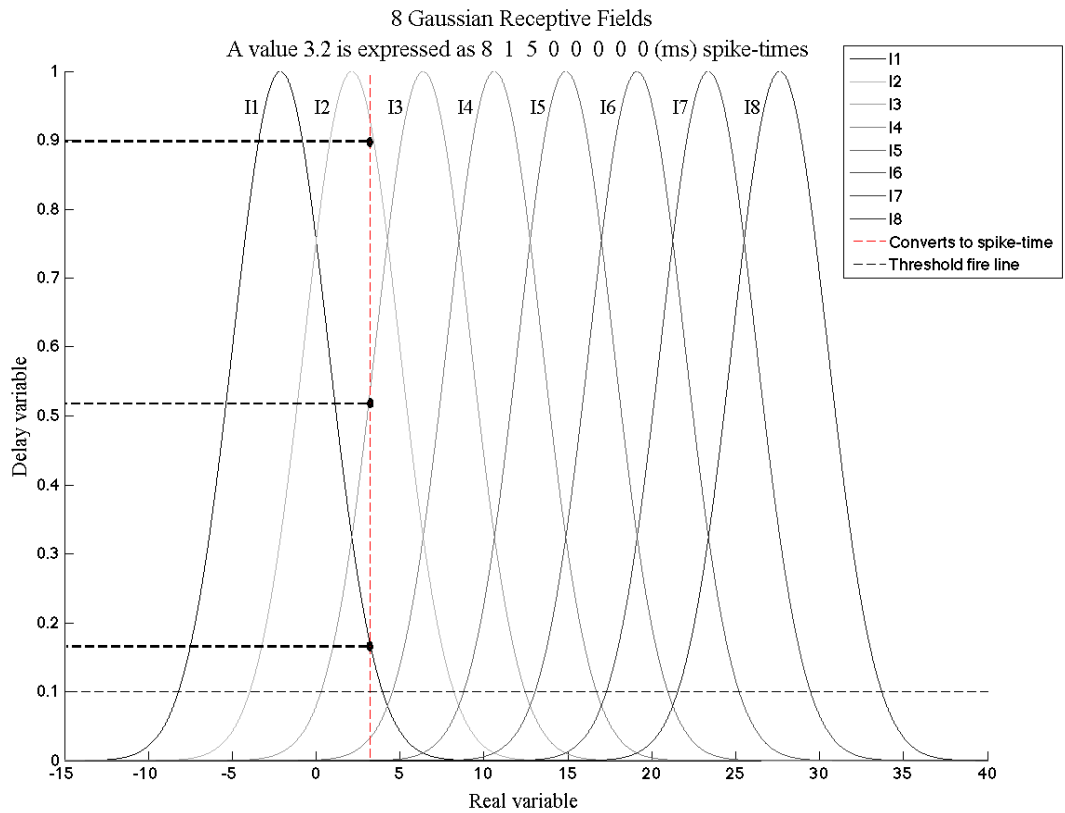

**Figure 5.2: Conversion of a real number, in this case 3.2 into spike-times using 8 Gaussian RF neurons.**

## 5.4. The proposed network architecture

The proposed network architecture has 3 layers and it can be seen in Figure 5.3. In the input layer there are 33 neurons because 8 Gaussian Receptive Field neurons are used for each iris feature and also 1 bias neuron is needed. In the hidden layer there are 8 neurons, as proposed by [43] and the output layer has only one neuron which emit a spike at 15ms to indicate the Setosa class, 20ms for the Versicolor class and at 25 for the Virginica class.

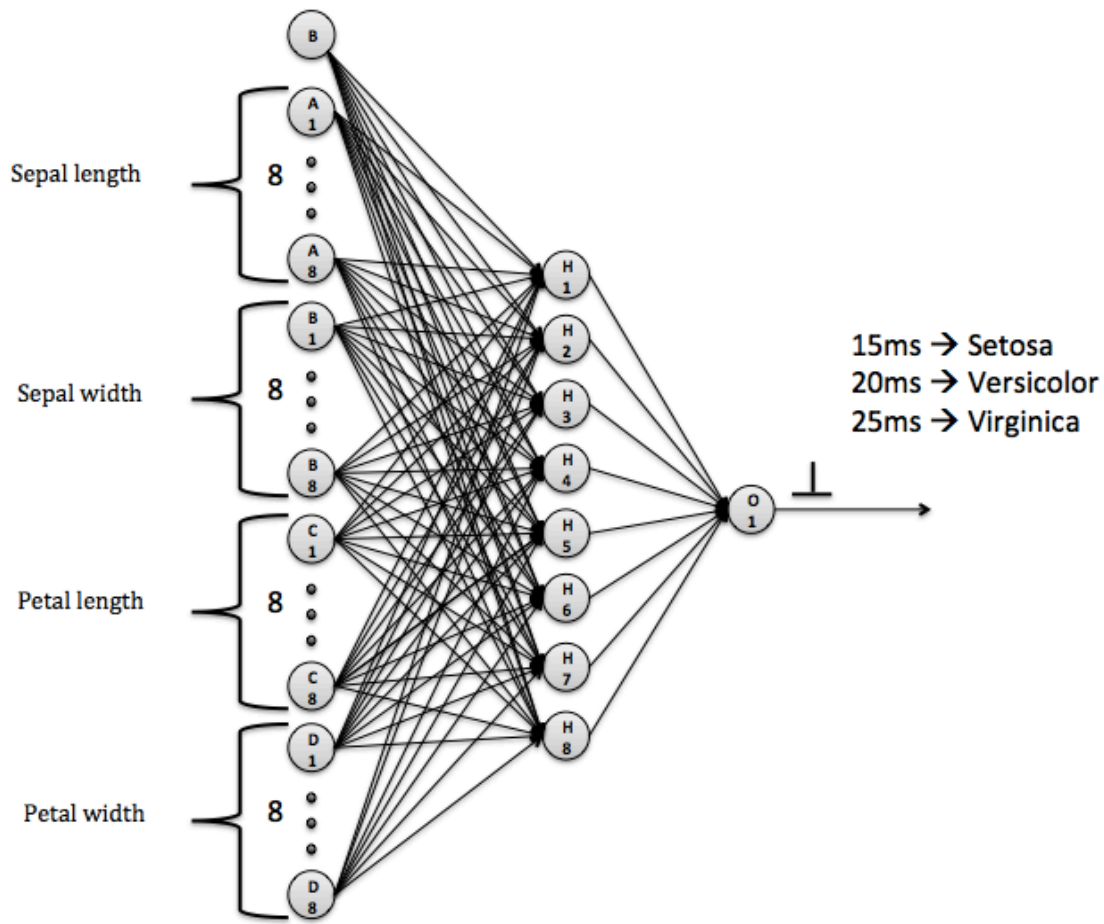

Figure 5.3: The proposed network architecture, Topology=[33 8 1].

## 5.5. The training sets, cross validation schemes and settings

The iris dataset has 150 input patterns in total so it was split into the following training sets:

- 30 training set, 10 input patterns from each class.
- 60 training set, 20 input patterns from each class.
- 75 training set, 25 input patterns from each class.
- 90 training set, 30 input patterns from each class.

The K-Fold cross validation scheme was used for performance estimation. In K-Fold cross validation the dataset is split into K subsets, also known as folds, and each time each K subset is used for the training process while the remaining subsets are used for validation. Finally, the mean error is estimated as follows:

$$E = \frac{1}{K}\sum_{k=1}^{K} E_{val}(k) \quad (5.1)$$

The advantage of K-fold cross validation is that all the input patterns in the dataset are eventually used for both training and testing. The K-fold cross validation schemes that were used can be observed in Figure 5.4.

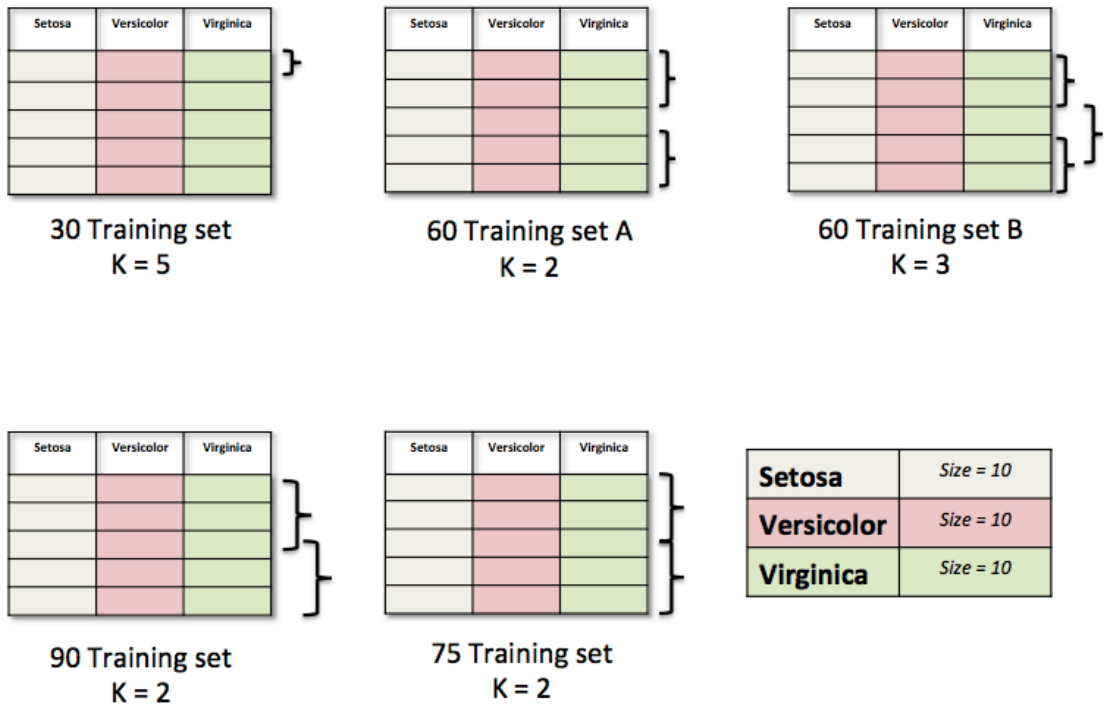

Figure 5.4: The 4 different training sets and the 5 K-Fold cross validation schemes that were used.

The spiking neural network settings can be seen in Table 5.2, while the genetic algorithm settings are in Table 5.3. Furthermore, two termination heuristics were set this time. The first one was when the mean squared error was below 0.25 and the second one was a maximum generation that the genetic algorithm was allowed to reach. The second heuristic is needed since the genetic algorithm is a stochastic method and a convergence is not always guaranteed.

Finally, in some cases the population size had to be increased since the proposed algorithm was not able to converge to the requested mean squared error, before reaching the maximum generation.

The hierarchy chart of the Fisher iris-training program can be observed in Appendix C.2, while the additional Matlab functions can be seen in Appendix B, from B.3.3 to B.3.7.

| Table 5.2: Spiking neural network settings | | Table 5.3: Genetic algorithm settings | |
|---|---|---|---|
| **Topology** | [33 8 1] | **Bits size of the weights** | 3 |
| **Simulation time** | 50ms | **Bits size of the delay times** | 3 |
| **Tau** | 3 | **Crossover rate** | 0.6 |
| **tauR** | 20 | **Mutation rate** | 0.01 |
| **Threshold for 1 binary decimal weights** | 3 | **Selective pressure** | 1.5 |
| | | **Elitism operator** | 8 |
| **Threshold for integer weights** | 6 | **Population size** | 600 |
| **Maximum spikes** | 10 | **Maximum generation** | 600 |
| **Time step** | 1 | | |

## 5.6. Results and discussion

An output spike is characterized as misclassified if it is more than 2ms away from the desired spike-time; the same method was used by Gosh-Dastidar et al. [43].

Furthermore, the high-throughput computer service, Condor [46], from Liverpool university was used since each training would last for a number of days. The iris training program was modified to allow checkpoints, so that if a training process was stopped, it would continue from the last saved generation and not from the start.

The results of the Fisher iris training, for each K-Fold of the five training sets, are summarized in Table 5.4. While in Table 5.5 the classification accuracy (C.A.) and the mean error for each K-Fold cross validation scheme can be seen.

Table 5.4: The number of misclassified output spikes for each training set and for each K-Fold.

| Training set | 1 binary decimal weights | | | | | Integer weights | | | | |
| --- | --- | --- | --- | --- | --- | --- | --- | --- | --- | --- |
| | K-Fold cross validation | | | | | K-Fold cross validation | | | | |
| | K1 | K2 | K3 | K4 | K5 | K1 | K2 | K3 | K4 | K5 |
| 30 | 16 | 12 | 19 | 12 | 5 | 15 | 12 | 19 | 12 | 5 |
| 60 A | 3 | 5 | 6 | | | 7 | 7 | 6 | | |
| 60 B | 3 | 6 | | | | 7 | 6 | | | |
| 75 | 3 | 9 | | | | 7 | 8 | | | |
| 90 | 2 | 8 | | | | 3 | 6 | | | |

Table 5.5: The mean error for each K-Fold cross validation scheme and the classification accuracy (C.A.).

| Training set | Folds | 1 binary decimal weights | | Integer weights | |
| --- | --- | --- | --- | --- | --- |
| | | Mean error | C.A. | Mean error | C.A. |
| 30 | 5-fold CV | 12.8 | 91.46% | 12.6 | 91.6% |
| 60 A | 3-Fold CV | 4.66 | 96.89% | 6.66 | 95.56% |
| 60 B | 2-Fold CV | 4.5 | 97% | 6.5 | 95.66% |
| 75 | 2-Fold CV | 6 | 96% | 7.5 | 95% |
| 90 | 2-Fold CV | 5 | 96.66% | 4.5 | 97% |

Finally, in Table 5.6 there is a comparison between the proposed algorithm (P.A.) and the ones that exist in the literature. As can be seen, even though 3 bits were used for the synaptic weights and delay times, the proposed algorithm produced better results compared to the real-valued weights of the SpikeProp, QuickProp and Rprop, for the same training sets [43].

Table 5.6: A comparison of the classification accuracy between the SpikeProp, QuickProp, Rprop [43] and the proposed algorithm (P.A.).

| Training set | SpikeProp | QuickProp | Rprop | P.A. with 1 binary decimal weights | P.A. with integer weights |
| --- | --- | --- | --- | --- | --- |
| 30 | 92.7% | 85.2% | 90.3% | 91.46% | 91.6% |
| 60 A | 91.9% | 91% | 94.8% | 96.89% | 95.56% |
| 60 B | 91.9% | 91% | 94.8% | 97% | 95.66% |
| 75 | 85.2% | 92.3% | 93.2% | 96% | 95% |
| 90 | 86.2% | 91.7% | 93.5% | 96.66% | 97% |

For the following training sets, the population size of the genetic algorithm had to be increased from 600 to 1000 in order to achieve the desirable mean squared error: the training set 90 and from the training set 60 A, the 2$^{nd}$ fold (K=2).

The genetic algorithm plots can be found in Appendix E.2, while the histograms can be seen in Appendix G. Finally, the misclassified spikes for each training set and each fold can be observed in Appendix F. The tables of the trained synaptic weights and delays were not included due to their size. However, they can be exported to Excel using the GUI program as described in Appendix M.

# 6. Hardware implementation

A hardware implementation was done, in order to prove that the trained synaptic weights and delays could be imported to a hardware system and produce comparable results to the simulations. A hardware simulation was done first, so that the behaviour of the system, on real hardware could be seen. This was needed in order to see the difference between the virtual-time of Matlab environment and the real-time of the processor speed that the spiking neuron was going to be implemented.

## 6.1. Simulation of the one neuron XOR in a hardware processor

The simulation process was done on Simulink using the Stateflow library. The spiking neuron was designed using the flowchart that can be seen in Figures 6.2 & 6.3 and the Data Table of the program can be seen in Appendix L.  The one neuron XOR with integer weights was implemented. A 10MHz clock was used as processor clock for the Stateflow system. The whole system can be seen in Figure 6.1.

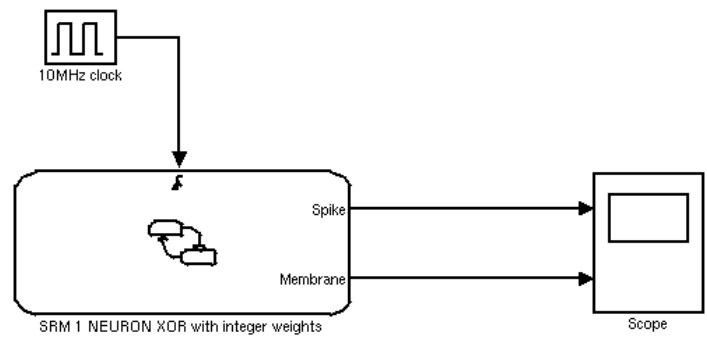

Figure 6.1: The one neuron XOR with integer weights hardware simulation

Two simplifications were made. The first one was that each statement would need one clock cycle to execute and the second one was that instead of receiving input spikes as digital input signals the input spike patterns were saved as counter values. Both of these simplifications do not affect the results.

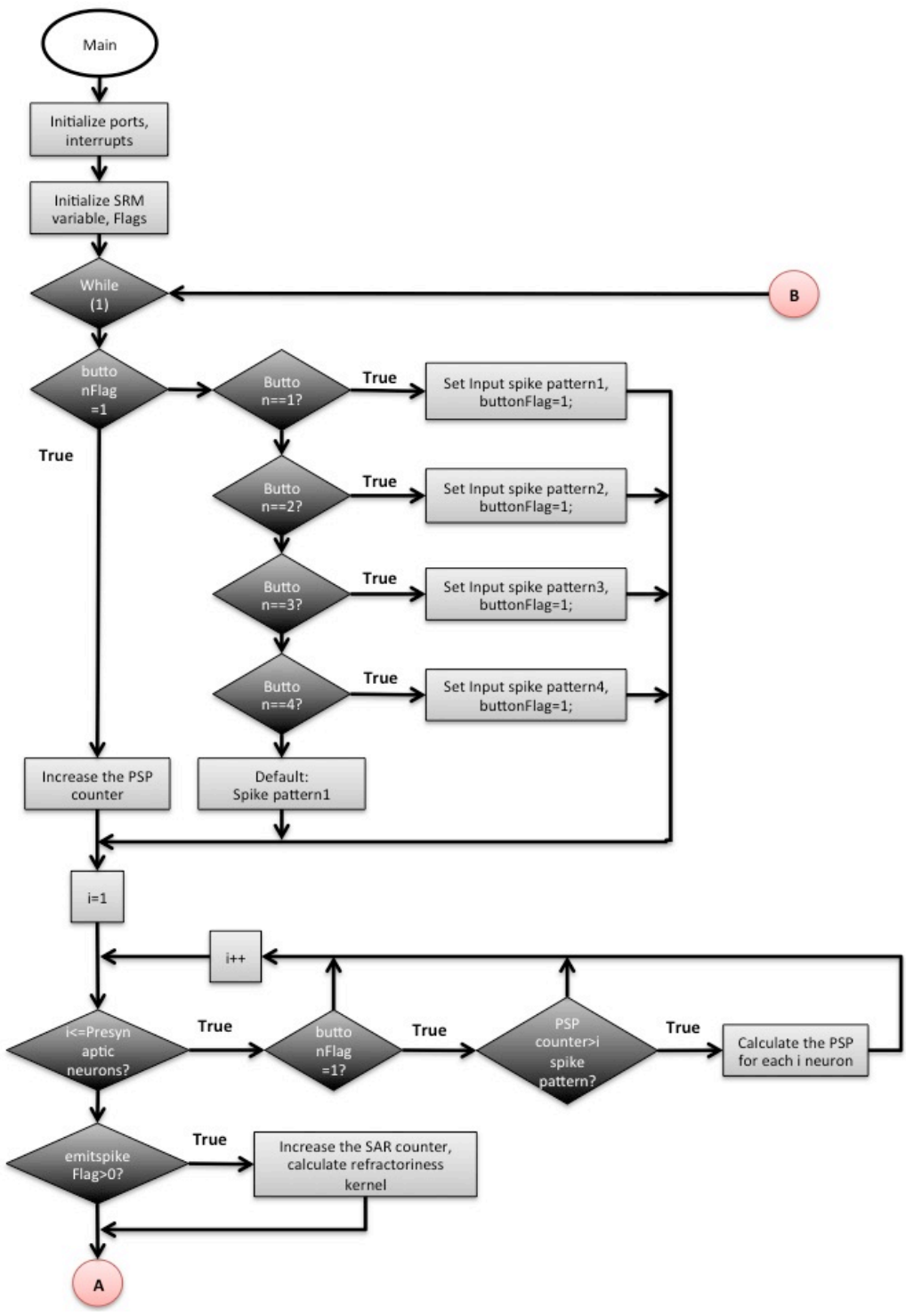

Figure 6.2: The flowchart of the one neuron XOR program. Part 1 of 2.

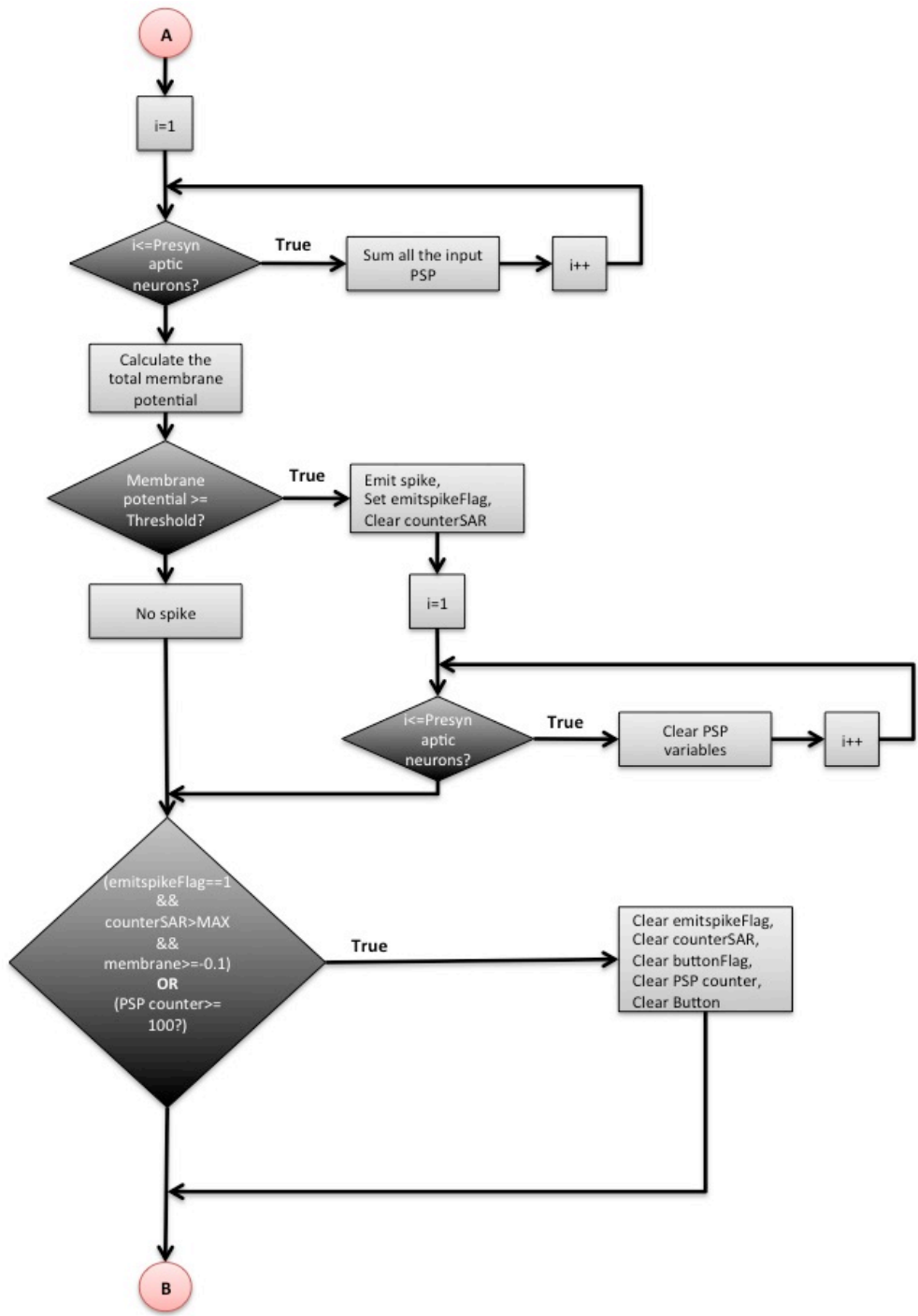

**Figure 6.3: The flowchart of the one neuron XOR program. Part 2 of 2.**

The trained synaptic weights and delays that were used for this simulation can be seen in Table 4.25. Finally, the same spiking neuron settings were used as in Table 4.23. The simulation results can be seen in the following section.

### 6.2. Simulation results and discussion

As stated before, the input spike patterns and the output spike times that were expressed as milliseconds in the Matlab virtual-time simulations, now, are expressed as values of the main software counter, counterMAIN from Appendix L.

The Simulink plots can be seen in Figure 6.4 and if compared to the ones from Figure 4.21 it can be seen that the results are similar. The only difference is that the output spike value of the counterMAIN should be 12 instead of 13 for the input spike pattern: 117 and 10 instead of 11 for the input spike pattern: 171. This difference is because of the different software implementation, the main software counter counterMAIN starts counting as soon as an input spike pattern has been selected prior to any computation process of the postsynaptic potential takes place, Figure 6.2.

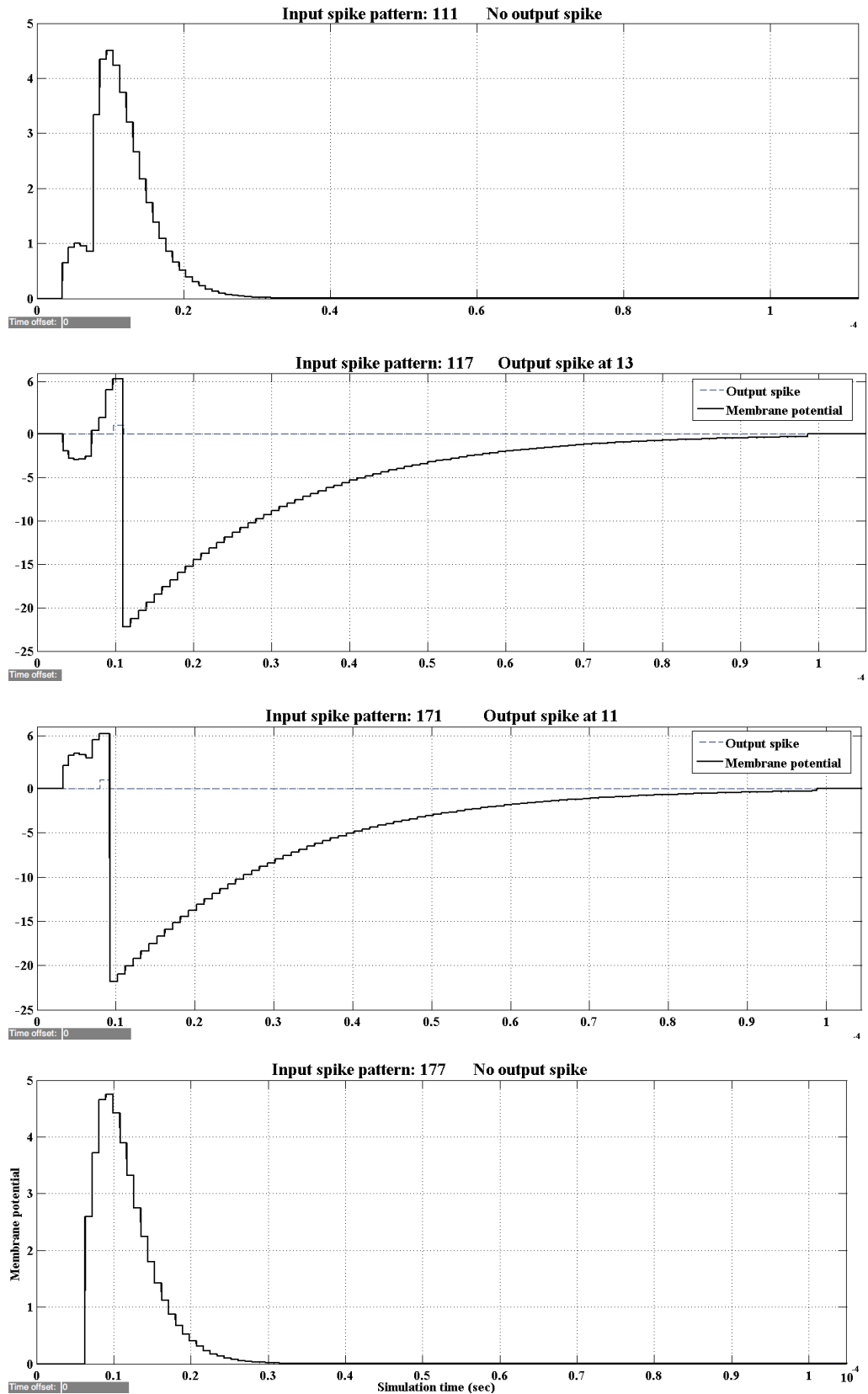

**Figure 6.4: The Simulink plots of the one neuron XOR its four input patterns.**

## 6.3. Running the one neuron XOR on the Stellaris LM3S6965 evaluation kit

The development board that was used was the Stellaris LM3S6965 Evaluation Kit and it can be seen in Figure 6.5.

The Stellaris LM3S6965 Evaluation Kit features [48]:

- LM3S6965 Evaluation Board
- Stellaris LM3S6965 microcontroller which has an ARM Cortex-M3 processor and a fully integrated 10/100 (MAC+PHY) Ethernet controller
- Simple setup: USB cable provides serial communication, debugging, and power
- OLED graphics display with 128 x 64 pixel resolution and 16 shades of gray
- User LED, navigation switches, and select pushbuttons
- Magnetic speaker
- All LM3S6965 I/O available on labeled break-out pads
- Standard ARM® 20-pin JTAG debug connector with input and output modes
- MicroSD card slot
- Retracting Ethernet cable, USB cable, and JTAG cable

For this project, the navigation switches were used for the selection of the input spike patterns; the OLED graphics display was used to display the main selection menu and the results. Finally, the status LED was used to indicate a UART transmission; the LED turns on when there is an ongoing transmission. The selection menu can be seen in Figure 6.5.

The one neuron XOR program for the LM3S6965 was written in C programming language using the Keil's μVision 4 IDE and compiler. The source code and start-up code can be seen in Appendix B.4, while the Data Table can be seen in Appendix L. Same as before, the trained synaptic weights and delays that were used can be seen in Table 4.25 and the spiking neuron settings can be seen in Table 4.23. The clock of the microcontroller was set to 8MHz.

The user chooses which input pattern wants to run from the four navigation switches. The "up" button runs the 1st input spike pattern: {111}, the "left" button runs the 2nd input spike pattern: {117}, the "right" button runs the 3rd input spike pattern: {171} and finally the "down" button runs the 4th input spike pattern: {177}.

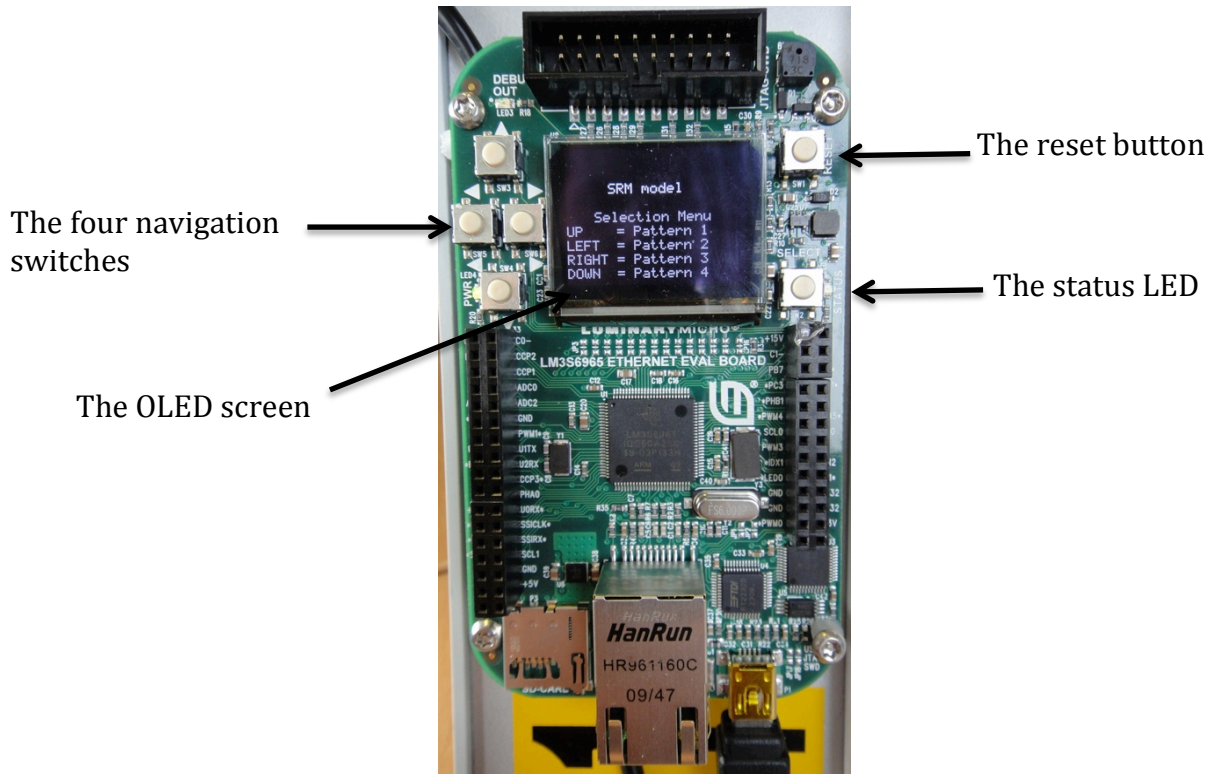

**Figure 6.5: The main selection menu of the one neuron XOR hardware implementation on the LM3S6965 evaluation board.**

### 6.4. Hardware results and discussion

The ARM Cortex-M3 SysTick timer was used in order to calculate the total clock cycles needed for the LM3S6965 to emit an output spike. The results are summarized in Table 6.1, while the plots of the received date, from the LM3S6965 UART port, can be seen in Figure 6.6. For more information on the received data please refer to Appendix H.

Table 6.1: The time and the total clock cycles the embedded system needs to emit a spike.

|  | For a 8MHz processor speed | For a 50MHz processor speed | Total clock cycles |
|---|---|---|---|
| **Input spike pattern 1** | - | - | - |
| **Input spike pattern 2** | 2.53ms | 404.64μs | 20232 |
| **Input spike pattern 3** | 1.95ms | 312.32μs | 15616 |
| **Input spike pattern 4** | - | - | - |

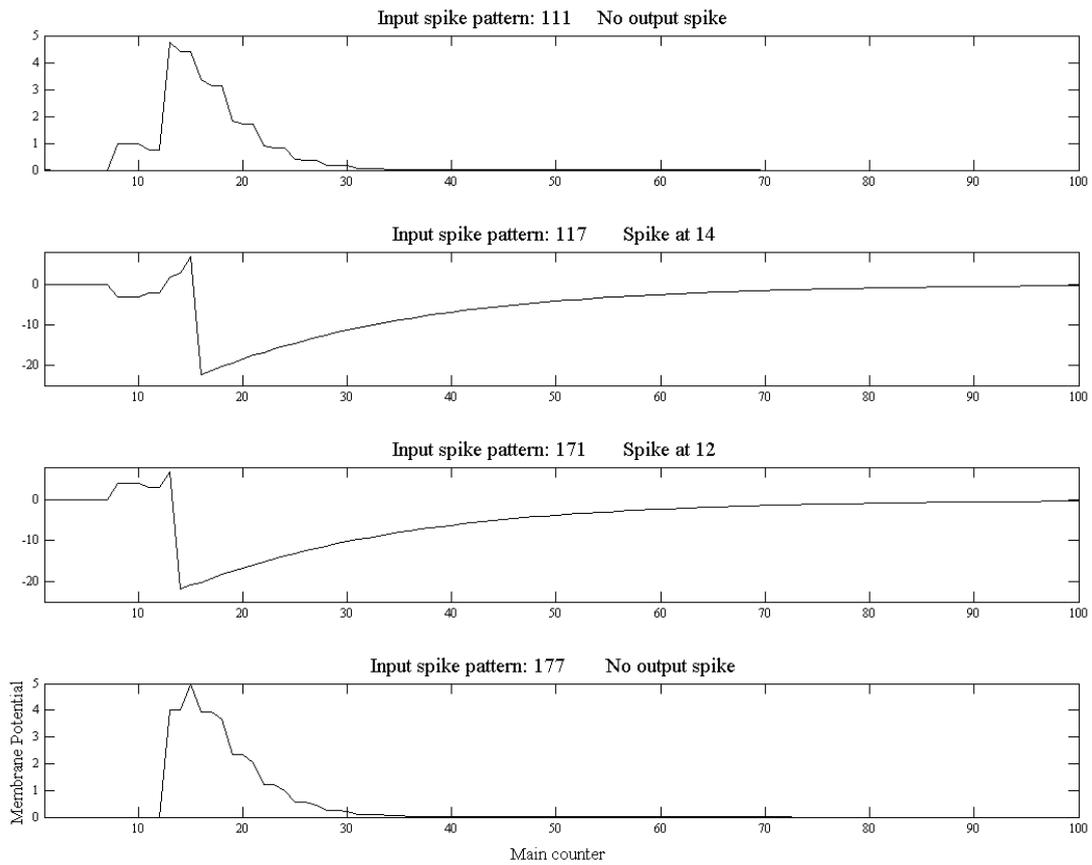

Figure 6.6: The membrane potential of the one neuron XOR, for the four input spike patterns, on the Stellaris LM3S6965 evaluation kit.

As it can be seen in Figure 6.6 and compare to Figure 6.4 the hardware implementation of the one neuron XOR shows the same behaviour. In Appendix I, the Matlab simulation, hardware simulation and hardware implementation plots are shown together, pattern by pattern, for comparison reasons.

The hardware implementation results proved that the trained synaptic weights and delay times of the Matlab program could be imported to a hardware system and produce similar results at fast speeds.

# 7. Conclusions and future work

## 7.1. Conclusions

In this project a new supervised training algorithm for spiking neural networks was developed using Genetic Algorithms. The proposed algorithm was designed for singe weights and delay times per synapse, in contrast to the SpikeProp, QuickProp and Rprop that operate on networks with multiple sub-synapses per neuron. Single synapse per neuron means fewer calculations since less postsynaptic potentials for each neuron would have to be calculated.

In addition, the proposed algorithm was able to train both synaptic weights and delays in contrast to the aforementioned algorithms that only train the weights. Even thought there are a numerous training algorithms based on Evolutionary Algorithms that also train both weights and delay times, the proposed one uses limited precision representation for the weights and delay values. Furthermore, multiple spikes per neuron are allowed taking full advantage of the computational powers that the spiking neurons have.

Two different limited precision schemes were studied in this project. The first one was the 3 bit integer delay times and the 3 bit with one binary decimal place weights and the second scheme used 3 bit integer values for the delays and 3 bits integer values for the weights. In both cases only six bits were used to describe a synapse.

In the case of the XOR problem, the proposed algorithm was able to converge to a much smaller mean squared error for the same network architecture compared to SpikeProp, QuickProp and Rprop. In addition, it was able to converge for a smaller network than the one proposed. Finally, it was able to solve the XOR problem using only one neuron, something that is impossible for the traditional artificial neural networks, proving the power of processing information in a spatial-temporal way.

Furthermore, for the Fisher iris classification problem and for the same training sets, the proposed algorithm produced higher classification accuracy when compared to the aforementioned training algorithms.

Finally, a simple hardware implementation was done on a microcontroller that used an ARM Cortex-M3 processor for the one neuron XOR in order to prove that the trained Matlab data can be imported to the system and produce similar results.

The main disadvantage of the proposed algorithm is the time needed for a training, which increases with the increase of the population size, training set or when a smaller time step is chosen.

## 7.2. Future work

To conclude, biological neural networks are fault tolerant and operate in a massively parallel way. Exploiting these features, processors could be designed using CMOS technology to create spiking neural networks. Then, the synaptic weights and delay times could be programmed to a specific task.

The advantage of this design architecture is that if one neuron is damaged, the ability of the processor would degrade but it would continue to operate in contrast to conventional computers, where if one transistor stops working, results in total failure of the system.

Furthermore, these processors would offer very fast processing speeds and better power efficiencies since neurons operate in parallel and in an asynchronous mode. These processors could be used in pattern recognition, robotics and machine vision tasks.

IBM has recently designed a processor like the aforementioned one, using 45nm SOI-CMOS technology. However, in that processor only the synaptic weights are programmable. The project is named "SyNAPSE" and the funding received was 21 million dollars from the Defense Advanced Research Projects Agency (DARPA) [49].

### 7.2.1. Regarding the proposed algorithm

The proposed Genetic Algorithm for the crossover operation used the uniform crossover in order not to mix synaptic weights and delays with each other and from different layers of the neural network. Nevertheless, it would be interesting to compare these results to the classical x-point crossover scheme.

Also, more bits could be used for the representation of the synaptic weights in order to increase the resolution of the decimal place and see if lower mean squared errors are achievable.

Furthermore, limited precision in unsupervised clustering methods could be explored using the same learning rule as defined in references [11,25] and explained in section 1.5.1. It should be pointed out that the same spiking neuron model was used in the aforementioned works. Lastly, limited precision could be studied for some other spiking neuron models that are better for hardware implementation such as the Izhikevich and the Leaky Integrate & Fire models.

**7.2.2. Regarding the training speed**

One of the biggest disadvantages of the proposed algorithm is the training time. One way to overcome this problem would be to implement spiking neurons on hardware and connect them together in order to perform the training process on the hardware. The Matlab program could be used to send the input spikes, then receive the output spikes, calculate the objective function for each individual and finally update the synaptic weights and delay times.

The hardware implementation of the spiking neural network could be either done on a FPGA board or microcontrollers. The advantages of the FPGA implementation are the processing speed of the neurons and size, since the neural network is on one chip. However, the advantages of implementing the spiking neurons on microcontrollers and using C or Assembly programming language are that the connections between the neurons are not physical, as in the case of the FPGA implementation, but logical, which means new connections could be done easily. The disadvantages in the latter case are the total size and processing speed when compared to the FPGA implementation.

The LM3S6965 evaluation board that was used for this project includes an Ethernet controller, so if a board would represent a spiking neuron then all the boards could be connected to an Ethernet hub and then neurons could send a spike to each other using I.P. addresses. A similar approach to this has been done on the SpiNNaker project, in the University of Manchester, using the Address-Event Representation (AER) to transmit spikes between the ARM processors [50]. Additionally, using this method would allow further research in optimizing the neural network architecture to a given problem using Genetic Programming.

### 7.2.3. Regarding the limited precision schemes

In this thesis, only the synaptic weights and delays were expressed with limited precision. The membrane potential, postsynaptic and refractoriness kernels of the SRM used double or single precision representation. It would be worth investigating on what happens when fixed-point representation is used instead. This along with the limited precision synaptic weights and delays could result in less memory for each neuron, which means designs with smaller size, lower cost and complexity. In the end, how much precision is really needed?

# A. Project Specification

**Student Name:** Evangelos Stromatias      **Project Code:** 32

**E-mail:** evangelos.stromatias@liv.ac.uk    **Supervisor:** Dr. John Marsland

**Project Title:**    Developing a supervised training algorithm for limited precision feed forward spiking neural networks

## Project Statement

**Project objective:**

Spiking neural networks have been referred to as the third generation of artificial neural networks and even though they are more powerful than the traditional artificial neural networks [1] they still have not gained the success that their predecessors did. This is due to the fact that a supervised training algorithm similar to the error backpropagation cannot be directly implemented due to their threshold function.

**Overall objective:**

Spiking neural networks are more suitable for parallel processing since each neuron operate asynchronously to each other. An important factor when it comes to hardware implementation of spiking neural networks is the precision of the weights since they can only be represented by a finite number of bits. Thus reducing their precision reduces the total cost and size. Furthermore the available supervised training algorithms based on the backpropagation, like Spikeprop [2-3] use a lot of assumptions and limitations. A supervised training algorithm that trains both weights and delay times of the synapses with limited precision would be very useful for future hardware implementations of spiking neural networks.

**Project methodology:**

Firstly, a Matlab program has to be developed in order to create spiking neural networks with different architectures and observe the spikes as they propagate. Then a supervised training algorithm will be created based on the Evolutionary Algorithms in order to train both the weights and the delay times of the synapses. Two decoding schemes will be investigated: 3bit integer weight values and 3bit with one binary decimal weight values. The proposed algorithm will be benchmarked on the XOR classification problem and the Fisher iris dataset. Then a hardware simulation will be done using the Stateflow library. Finally a spiking neural network will be implemented on the Luminary Stellaris LM3S6965 evaluation board [4], using the trained data from the Matlab program.

## B. Outline of the workplan

The first step of this project is the development of a Matlab program that will create feed forward spiking neural networks, for any architecture, using the Spike Response Model (SRM); this program will later be used as a function in the training. The proposed supervised training algorithm will be created based on Genetic Algorithms and will be able to train the weights and the synapse delay times of a spiking neural network with limited precision. Then the training algorithm will be tested on two different decoding schemes in three classification problems: the XOR and the Fisher iris dataset. For the latter one, the K-fold cross validation method will be used in order to calculate the actual error. After that a hardware simulation will be done using the Stateflow library in Simulink/Matlab. The hardware simulation part is necessary in order to observe how the neurons behave in real-time (e.g. based on the microprocessor speed) in contrast to the Matlab virtual training time. Finally, a hardware implementation of the XOR problem will be made on the Luminary Stellaris LM3S6965 evaluation board, using the trained data from the Matlab program for integer weights.

| TASKS                                                                                             WEEK No | 1 | 2 | 3 | 4 | 5 | 6 | 7 | 8 | 9 | 10 | 11 | 12 |
|---|---|---|---|---|---|---|---|---|---|---|---|---|
| Creating the Spike Response Model $SRM_0$ | ■ | ◇1 | | | | | | | | | | |
| Matlab program for spiking neural networks | ■ | ■ | | | | | | | | | | |
| Creating a supervised training algorithm based on Genetic Algorithms with limited precision | | | ■ | | | | | | | | | |
| Training the Spiking neural network for the XOR problem with 3bit decimal binary weights | | | | ■ | | ◇2 | | | | | | |
| Training the Spiking neural network for the XOR problem with 3bit integer weights | | | | | ■ | | | | | | | |
| Interim Report | | | | | ■ | ■ | | | | ◇3 | | |
| Training the Spiking neural network for the iris classification problem | | | | | | | ■ | ■ | ■ | | | |
| Simulation of the hardware implementation | | | | | | | | | ■ | ◇4 | | |
| Hardware implementation (XOR problem) | | | | | | | | | | ■ | ◇5 | |
| Experimental results & conclusions | | | | | | | | | | | ■ | ◇6 |
| Thesis writing | | | | | | | | | | ■ | ■ | ■ |

Milestones:
1. A spiking neural network function has been created.
2. Results for the XOR classification problem.
3. Results for the iris classification problem.
4. Results for the hardware simulation.
5. Experimental results.
6. Project completed.

**The rest of the appendices can be accessed through the attached CD.**